\documentclass[a4paper,fleqn]{cas-sc}

\def\tsc#1{\csdef{#1}{\textsc{\lowercase{#1}}\xspace}}
\tsc{WGM}
\tsc{QE}

\usepackage{caption}
\usepackage[noabbrev,capitalize,nameinlink]{cleveref}
\usepackage{amssymb}
\usepackage{makecell}
\usepackage{tabularx}
\usepackage{fancyhdr}
\usepackage{float}
\usepackage{enumitem}
\usepackage{pifont} 
\usepackage{xcolor}
\usepackage{soul}
\usepackage[authoryear]{natbib}
\usepackage{booktabs}
\usepackage{lineno}

\begin{document}
\let\WriteBookmarks\relax
\def\floatpagepagefraction{1}
\def\textpagefraction{.001}

\shorttitle{WSF-MI for Large-scale Cropland Mapping}    

\shortauthors{Yuze.W}  



\title [mode = title]{Weakly Supervised Framework Considering Multi-temporal Information for Large-scale Cropland Mapping with Satellite Imagery}  

\tnotemark[1] 

\tnotetext[1]{} 

%

\author[1]{Yuze Wang}
\fnmark[1]
\credit{Conceptualization, Data curation, Formal analysis, Investigation, Resources, Software, Validation, Visualization, Writing – original draft, Writing – review \& editing}

\author[1]{Aoran Hu}
\credit{Conceptualization, Data curation, Methodology, Resources, Software, Writing – original draft}

\author[1]{Ji Qi}
\credit{Investigation, Methodology, Resources, Visualization}

\author[1,2]{Yang Liu}
\credit{Funding acquisition, Project administration, Supervision}

\author[1]{Chao Tao}[orcid = 0000-0003-0071-310X]
\cortext[1]{Chao Tao}
\ead[url]{kingtaochao@126.com}
\cormark[1]
\credit{Funding acquisition, Project administration, Resources, Supervision, Writing – original draft, Writing – review \& editing}





\affiliation[1]{organization={School of Geosciences and Info-Physics, Central South University},
            addressline={No. 932, Lushan Nan Road}, 
            city={Changsha},
            postcode={410083}, 
            country={China}}
\affiliation[1]{organization={The 27th Research Institute, China Electronic Technology Group Corporation},
            city={Zhengzhou},
            postcode={450047}, 
            country={China}}










\begin{abstract}
Accurately mapping large-scale cropland is crucial for agricultural production management and planning. Currently, the combination of remote sensing data and deep learning techniques has shown outstanding performance in cropland mapping. However, those approaches require massive precise labels, which are labor-intensive. To reduce the label cost, this study presented a weakly supervised framework considering multi-temporal information for large-scale cropland mapping. Specifically, we extract high-quality labels according to their consistency among global land cover (GLC) products to construct the supervised learning signal. On the one hand, to alleviate the over-fitting problem caused by the model’s over-trust of remaining errors in high-quality labels, we encode the similarity/aggregation of cropland in the visual/spatial domain to construct the unsupervised learning signal, and take it as the regularization term to constrain the supervised part. On the other hand, to sufficiently leverage the plentiful information in the samples without high-quality labels, we also incorporate the unsupervised learning signal in these samples, enriching the diversity of the feature space. After that, to capture the phenological features of croplands, we introduce dense satellite image time series (SITS) to extend the proposed framework in the temporal dimension. We also visualized the high-dimensional phenological features to uncover how multi-temporal information benefits cropland extraction, and assessed the method's robustness under conditions of data scarcity. The proposed framework has been experimentally validated for strong adaptability across three study areas (Hunan Province, Southeast France, and Kansas) in large-scale cropland mapping, and the internal mechanism and temporal generalizability are also investigated. The source codes are available at \url{https://github.com/wangyuze-csu/WSFCMI}.
\end{abstract}


\begin{highlights}
\item 
A weakly-supervised framework based on SITS for large-scale cropland mapping.
\item 
Encode intrinsic features to optimize the utilization of labels from GLC products
\item 
Experiments on three agricultural areas showed the advantage of the proposed method.
\item 
Uncover the benefits of using multi-temporal information in cropland extraction.
\item 
The methods exhibit robustness in data deficiency scenarios.
\end{highlights}

\begin{keywords}
Weakly supervised\sep Multi-temporal information\sep Large-scale cropland mapping\sep
\end{keywords}

\maketitle

\section{Introduction}\label{Section 1}

\par
Over the past decades, remote sensing observation has played significant roles in large-scale cropland mapping and monitoring \citep{huangAgriculturalRemoteSensing2018a,defournyRealtimeAgricultureMonitoring2019}. By offering timely and comprehensive images of nearly every part of the Earth's surface \citep{chiBigDataRemote2016a}, it served as a reliable information source for identifying cropland spatial distribution on a large scale \citep{weissRemoteSensingAgricultural2020a}. Moreover, it provides valuable support for various agricultural applications, such as land-use planning \citep{yinIntegratingRemoteSensing2021b}, food security \citep{karthikeyanReviewRemoteSensing2020a,calvaoRemoteSensingFood2015a}, and sustainable agroecology \citep{princeChallengesRemoteSensing2019a}. \par

With the accumulation of remote sensing data, several data-driven methods based on machine learning have been widely applied to the large-scale cropland mapping  \citep{zanagaESAWorldCover102022,donascimentobendiniDetailedAgriculturalLand2019a,pelletierAssessingRobustnessRandom2016a,gongFinerResolutionObservation2013a,zanagaESAWorldCover102022,yueTextureExtractionObjectoriented2013,northBoundaryDelineationAgricultural2019b,belgiuSentinel2CroplandMapping2018d,xuMonitoringCroplandChanges2017a}. The widely adopted approaches are employing traditional machine learning methods to analyze and interpret the remote sensing (RS) images. These methods often rely on a combination of low-level and middle-level visual features, such as texture \citep{yueTextureExtractionObjectoriented2013}, spectral \citep{northBoundaryDelineationAgricultural2019b}, and shape \citep{wagnerExtractingAgriculturalFields2020a} features. However, due to the influence of various factors on cropland, such as climate, geography, and topography, the methods that rely on handcrafted features typically encounter issues of restricted generalization performance and low accuracy \citep{nanniHandcraftedVsNonhandcrafted2017a}.In recent years, Deep Convolutional Neural Networks (DCNNs) \citep{lecunDeepLearning2015a} attracted great attention in cropland mapping \citep{singhDeepLearningMapping2022a,brownDynamicWorldRealtime2022b,zhangGeneralizedApproachBased2020e,sunUsingLongShortterm2019a,karraGlobalLandUse2021a,perselloDelineationAgriculturalFields2019}, because it can extract high-level visual features that are more representative and distinguishable. However, the performance of DCNNs shows a strong positive correlation with the number and diversity of high-quality labeled samples, leading to high labeling costs \citep{zhuDeepLearningRemote2017,liDeepNeuralNetwork2019a}. Although various methods \citep{hua2021semantic,lenczner2022dial} designed for sparse labeling conditions can significantly decrease the demand for labeling, the remaining label requirement in large-scale cropland mapping still implies a considerable manual cost. So, reducing the labeling cost while maintaining cropland mapping accuracy is still a great challenge.
\par

Some methods use existing global land cover (GLC) products, such as GFSAD 30 \citep{oliphantMappingCroplandExtent2019}, CCI-LC \citep{copernicusclimatechangeserviceLandCoverClassification2019a}, FROMGLC \citep{yuImproving30Global2013a}, MCD12Q1 \citep{friedlMCD12Q1MODISTerra2019}, Esri \citep{karraGlobalLandUse2021a}, ESA \citep{zanagaESAWorldCover102022}, DyWorld \citep{brownDynamicWorldRealtime2022b} as reference to train the model at a low cost and obtain accurate cropland mapping results \citep{liNovelAutomaticPhenology2021b,zhuOptimizingSelectionTraining2016a}. Those methods are known as Automatic Training Sample Generation (ATSG) \citep{liuNovelImperviousSurface2022a}. However, the label obtained from GLC products inevitably contains some errors, due to factors like the diversity of cropland scenes, imaging conditions, and classifier performance. Directly using those labels may lead to instability and over-fitting in the model learning process, causing low-quality cropland mapping. Therefore, identifying the errors of the labels generated from GLC products and preventing their negative impacts on the model training process is one of the significant research topics of ATSG methods. \par
According to the post-processing ways of error labels, the ATSG methods based on existing GLC products are divided into two categories: discard and re-correct
\par

Discard methods establish quality criteria to rate the labels from the GLC products, and exclude low-quality labels before training. Based on the MCD12Q1 product, \citet{zhangUsing500MODIS2017b} utilized temporal invariance as a quality criterion to discard the labels that have changes within three years, and excluded the labels with low classification confidence according to the quality assessment layer provided in the product auxiliary information. With the continuous release of GLC products, \citet{liCroplandDataFusion2020b} identified high-quality labels by considering their consistent performance across four GLC products (GFSAD 30, CCI-LC, FROMGLC, and MCD12Q1), and discarded the outlier labels based on the spectral distribution of all corresponding pixels. Considering the spectral mixing problem among vegetation cover, \citet{hermosillaLandCoverClassification2022a} integrated the 3D information from LiDAR data into the quality assessment of the labels from three products (NFI photo plot \citep{stinsonCanada2016}, EOSD \citep{wulderOperationalMappingLand2003} and NWS \citep{wulderNationalAssessmentWetland2018}). By doing so, they could identify and remove the low-quality labels that mismatch the vegetation height.
\par 
Re-correct methods construct filter strategies to obtain high-quality labels based on criteria such as classification consistency across multiple products \citep{hermosillaLandCoverClassification2022a} and spectral consistency among the same land covers \citep{liCroplandDataFusion2020b}. High-quality labels are then obtained and used as references to correct the remaining labels. Considering the spatial continuity and texture consistency within the same land cover,  \citet{chenNovelWeaklySupervised2023b} took the sub-regions obtained by Simple Non-Iterative Clustering (SNIC) as the correction unit, and reclassified each unit through voting among filtered high-quality labels, thereby extending high-quality labels into regions with low-quality labels. \citet{zhangNovelKnowledgeDrivenAutomated2023a} considered the phenological attribute of vegetation cover, and used high-quality labels and multi-temporal images to train the classifier to correct low-quality labels. \citet{nabourehNationalScaleLand2023a} selected pure pixels with high-quality labels as reference, and rectified the remaining pixels containing low-quality labels according to their spectral distance from the selected pixels.

\par

The above methods still encountered difficulties in achieving high-precision cropland extraction, thus falling short of meeting the localized needs of agricultural applications. The main reasons are as follows: 

\begin{enumerate}[label={\arabic*)},left=0pt]
\item  Insufficient use of samples with low-quality labels: Discard methods can mitigate the model’s over-fitting problem on incorrect labels, they also eliminate the samples with diverse features but low-quality labels before training. Furthermore, Discard methods may lead to the imbalance of intra-class distribution for training samples \citep{wangSemiSupervisedSemanticSegmentation2022b}, so the model can only learn typical features and has limited generalization capability. For example, in cropland mapping, these methods often constrain the trained classifier to only extract continuous and large block plain croplands, making it challenging to recognize cropland in complex environments, such as those scattered across hills and mountains.

\item Over-trust the samples with high-quality labels: Although re-correct methods allow diverse samples to participate in the model training process, they potentially introduce a lot of label noise, misleading the model during the learning process. This occurs for two reasons. First, due to the inherent limitations of the products, the filtering strategies struggle to completely eliminate all errors in the selected high-quality labels. Second, re-correct methods use samples with high-quality labels as references to correct the samples with low-quality labels, which may further amplify the remaining label errors. 
\end{enumerate}  \par

In summary, discard methods can ensure the overall quality of training samples but may decrease the diversity of the feature space due to the insufficient use of samples with low-quality labels. On the other hand, re-correct methods effectively utilize samples with low-quality labels, but place excessive trust in high-quality labels, which can amplify label errors and mislead the model optimization process. 
\par
To balance these two issues, we proposed a weakly-supervised framework that uses the labels from existing GLC products for large-scale cropland mapping. Specifically, to avoid the model over-trusting high-quality labels, we encoded the intrinsic feature distribution of the image to construct the unsupervised part of the learning signal, which was then used to constrain the supervised part directly constructed by the high-quality labels. It can prompt the model to assess and question the reliability of the supervised part of the learning signal, avoiding over-fitting the remaining errors in high-quality labels. Meanwhile, we applied the unsupervised part of the learning signal to the samples with low-quality labels to use the information contained to enhance the diversity of the model’s feature space. Additionally, we enhanced this framework by extending it into the temporal dimension, allowing the model to fully extract the phenological feature and change patterns of croplands from the high-density Satellite Image Time Series (SITS). The contributions of this paper are as follows: 
\begin{enumerate}[label={\arabic*)},left=0pt]
\item Given the high costs of large-scale cropland mapping, we proposed a weakly-supervised framework that leverages existing GLC data without manual labels. It further alleviates the impact of residual errors in filtered high-quality labeled samples, while effectively utilizing the plentiful information contained in low-quality labeled samples.
\item In the framework, we flexibly incorporated multi-temporal DCNNs with SITS to capture the phenological features of croplands. We further visualized these high-dimensional features to uncover how multi-temporal information enhances cropland extraction, and assessed the method's robustness under conditions of data scarcity to validate its practical applicability.
\item We conducted experiments in three study areas to test the feasibility and stability of our method for large-scale cropland mapping. We also investigated the internal working mechanisms and temporal generalizability of the proposed framework. 
\end{enumerate}  \par

The remainder of this paper is organized as follows: \cref{Section 2} presents the materials and process employed, and \cref{Section 3} describes the workflow of the proposed framework and the detail of applied network architecture. \cref{Section 4} and \cref{Section 5} present experimental results and discussion, and \cref{Section 6} summarizes and concludes the paper.
\par


\begin{figure}[htp]
	\centering
		\includegraphics[width=0.8\linewidth]{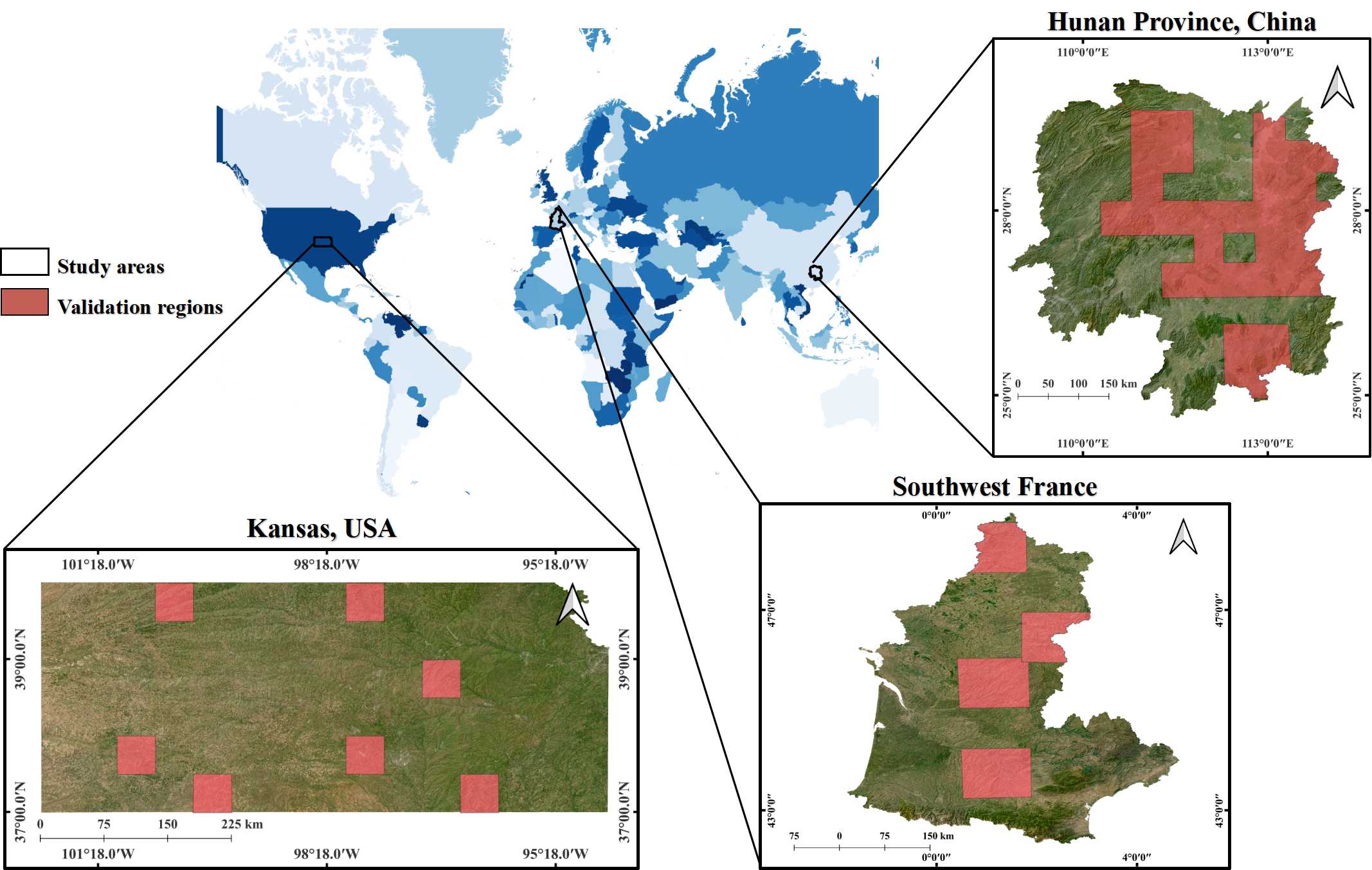}
	  \caption{The location and extent of the three study areas. The red regions are used to validate the accuracy of the cropland mapping result.}\label{fig1}
\end{figure}

\begin{table}[h] 
\caption{ The extent, climate, and main crops of the three study areas.}\label{tbl1}
{\scriptsize
\begin{tabular*}{\tblwidth}{@{}p{3cm}p{2cm}p{6cm}p{4.5cm}@{}}
\toprule
Study areas & Extent({km}$^2$) & Climate & Main crops \\ 
\midrule
Hunan Province,China & 211,800 & Continental subtropical monsoonal humid climate & Rice, Rapeseed, Cotton, Tea \\
Southwest France & 195,910 & Temperate maritime climate & Spring wheat, Soybeans, Olives\\
Kansas State, USA & 213,096 & Temperate continental climate & Winter wheat, Corn\\
\bottomrule
\end{tabular*}
}
\end{table}

\section{Materials and process}\label{Section 2}

\subsection{Study areas and satellite imagery}\label{Section 2.1}

\medskip

We chose three study areas across Asia, Europe, and North America, each representing distinct terrain landscapes, climatic regions, and agriculture systems, including different crop types and phenological attributes. Collectively, these regions encompass a vast area exceeding  620,806 {km}$^2$ (\cref{fig1}). The general information of the study areas is as follows (\cref{tbl1}) :

\begin{itemize}[left=0pt]
\item Hunan Province in China: Hunan province is situated in the central-southern part of China, covering a total area of 211,800 {km}$^2$. As one of China’s largest rice-planting bases, Hunan province ranks among the top ten in national grain production. The province’s diverse terrain, ranging from semi-alpine to low mountains, hills, basins, and plains, presents significant challenges for cropland monitoring. The area has a continental subtropical monsoonal humid climate, offering abundant light, heat, and water resources, which is suitable for cultivating a variety of crops including rice, rapeseed, cotton, and tea. The rice types grown here include double-season rice, medium-season rice, and late-season rice, each with distinct sowing and harvesting periods from April to November. 
\item Southwest France: We chose the southwest part of France, encompassing about 195,910 {km}$^2$, which represents 2/5 of the entire country, as our study area. This region has several major grain-producing areas such as Aquitaine Basin and Centre-Val de Loire. The landscape is dominated by basins and valleys. It has a temperate oceanic climate, suitable for cultivating a variety of crops, such as spring wheat, soybeans, and olives. Spring wheat and soybeans are typically sown in spring and harvested in late summer, while the olives are planted in early summer and harvested in mid-autumn. 
\item Kansas State in the USA: Kansas state is located in the middle of the USA, covering a total area of 213,096 {km}$^2$. It is a major grain-producing region in the United States, with the highest wheat production in the country. Kansas has large plains, suitable for extensive mechanized agricultural activities. The area has a temperate continental climate, favoring growing winter wheat and corn. Winter wheat is sown in mid-September and harvested in late June to early July of the following year. Corn is sown in mid-April and harvested in mid-October.
\end{itemize}  \par

We collected Sentinel-2A and Sentinel-2B Level-2A Bottom of Atmosphere reflectance images (S2 L2A) from Google Cloud, encompassing the three study areas for the period from January 2020 to December 2020. For cropland extraction, we selectively utilized the Blue, Green, Red, and Near-Infrared (NIR) bands (10 m spatial resolution), along with Narrow NIR, Red Edge (RE), and Short-Wave Infrared (SWIR) bands (20 m spatial resolution). Furthermore, we used the Quality Assurance (QA) bands provided by the ESA on Google Earth Engine (GEE) as a reference to select the images with cloud coverage less than 20\% \citep{amaniGoogleEarthEngine2020a}. The QA bands were also used to identify cloudy regions in the remaining images, which were filled using the cloud-free images from adjacent time phases. Finally, all SITS were composed of monthly images, where each month’s image was obtained by averaging all available images during the month.

\subsection{Global Cropland layer and Validation datasets}\label{Section 2.2}

\medskip
To generate training labels, we utilized the cropland layers from three GLC products with 10m spatial resolution: ESA World Cover, Eris Land Cover, and Dynamic World. Firstly, we established a definition of ’cropland’ by uniformly mapping from the related categories in various GLC products. Specifically, cropland is defined as fields that are covered by annual crops that are sown or planted and are capable of being harvested at least once within the 12 months after the date of sowing or planting. This type of annual cropland generates an herbaceous canopy, and may occasionally be intermixed with trees or shrubby vegetation \citep{FoodandAgricultureOrganizationoftheUnitedNations_SystemIntegratedAgricultural_2005}. Secondly, we collected their data through the GEE platform, and projected their geographic coordinates to the World Geodetic System 1984 (WGS 84), consistent with that of SITS. The details of the three GLC products are as follows:
\begin{itemize}[left=0pt]
\item  \textbf{ESA World Cover (ESA) }: It was developed as a part of the ESA WorldCover project, under the 5th Earth Observation Envelope Program (EOEP-5) of the European Space Agency. The product was generated by SITS from Sentinel-1 and Sentinel-2 with multiple random forest classifiers from 2020 to 2021. It contains 11 first-level categories, and we used the 'Cultivated areas' category from the 2020 product to obtain the labels. This category is defined as land covered with annual cropland that is sowed/planted and harvestable at least once within the 12 months after the sowing/planting date. The annual cropland produces an herbaceous cover and is sometimes combined with some tree or woody vegetation \citep{zanaga2022esa}.
\item  \textbf{Esri Land Cover (Esri) }: It is a 10-m resolution map of the Earth’s land surface, published by the Environmental Systems Research Institute. This map was annually generated from 2017 to 2022 using the composite Sentinel-2 satellite images by deep learning models. The product divides the land cover into 9 categories, and we used the 'cropland' category from the 2020 product to obtain the labels. This category includes crops, human-planted cereals, and other non-tree-height cultivated plants. \citep{karra2021global}.
\item  \textbf{Dynamic World (DyWorld) }: It was developed by Google and the World Resources Institute. It is a near real-time global land use and cover dataset, updated in sync with the revisit cycle of the Sentinel-2 satellite. The product was generated by deep learning models using available images from 2015 to 2023. We collated all available data from various time phases in 2020 and determined the land cover categories for each pixel based on the mode principle derived from the statistical analysis. It divides land cover into 9 categories, and we used the 'crop' category to obtain the labels. This category is defined as crops humans planted/plotted cereals. \citep{brown2022dynamic}. 
\end{itemize}  

To assess the accuracy and completeness of the cropland mapping obtained by the proposed method, we established validation datasets for the three study areas. Considering that random sampling alone may skew the validation datasets toward particular cropland types prevalent in the study areas, potentially compromising the comprehensiveness and objectivity of the cropland extraction assessment in diverse environments. To address this issue, we introduced manual intervention in the selection process for the validation area. Specifically, we classified the entire study area based on the topographical features, and then utilized this result as a foundation to manually refine the sub-areas that were initially selected through random sampling.

For the Hunan study area, which is predominantly covered by hills and mountains, we manually excluded certain mountainous and hilly sub-areas and included sub-areas dominated by plains. For the Kansas study area, where plains are the dominant landscape, we excluded some sub-areas of plains, and added sub-areas that include hills and mountains. The validation labels of Hunan and Kansas study areas were obtained by visually interpreting the RGB bands of Sentinel-2 time series images, assisted by high-resolution Google images. In the Southwest France study area, we utilized the S4A crop classification datasets \citep{sykasSentinel2MultiyearMulticountry2022b} as a basis and modified them for validation purposes. We initially checked all the labels, and selected sub-areas with relatively comprehensive cropland annotations for further supplementation and adjustment. Subsequently, we manually selected the sub-areas to ensure a balance among various types of cropland scenes. The sampling results of the final validation areas are shown in \cref{fig1}. 

\begin{figure}[t]
	\centering
		\includegraphics[width=0.85\linewidth]{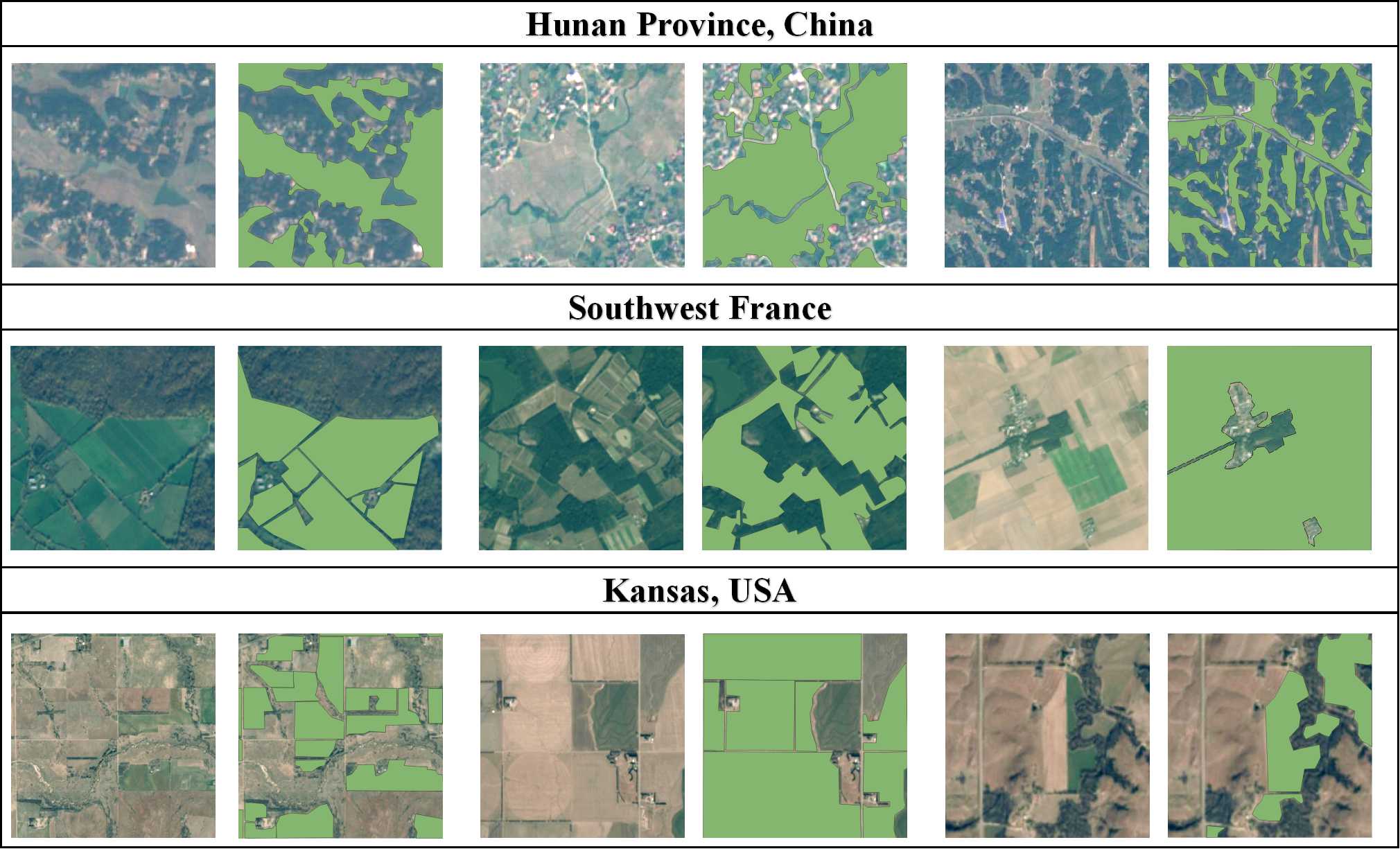}
	  \caption{ Examples of croplands in the study areas.}\label{fig2}
\end{figure}

Finally, in the Hunan study area, we labeled a total of 978,388 cropland fields, accounting for 18.00\% of the validation area. In the Southwest France study area, we identified 520,177 cropland fields, which covered 40.48\% of the validation area. In the Kansas study area, we identified 185,250 cropland fields, accounting for 48.24\% of the validation area. Note that, ’cropland fields’ refer to a piece of cropland separated from one another by identifiable boundaries \citep{fao2010world,xu2024deep}. Each image-level sample may include several cropland fields. Some typical samples of each study area are shown in \cref{fig2}.

\section{Methodology}\label{Section 3}
\begin{figure}[t]
	\centering
		\includegraphics[width=1\linewidth]{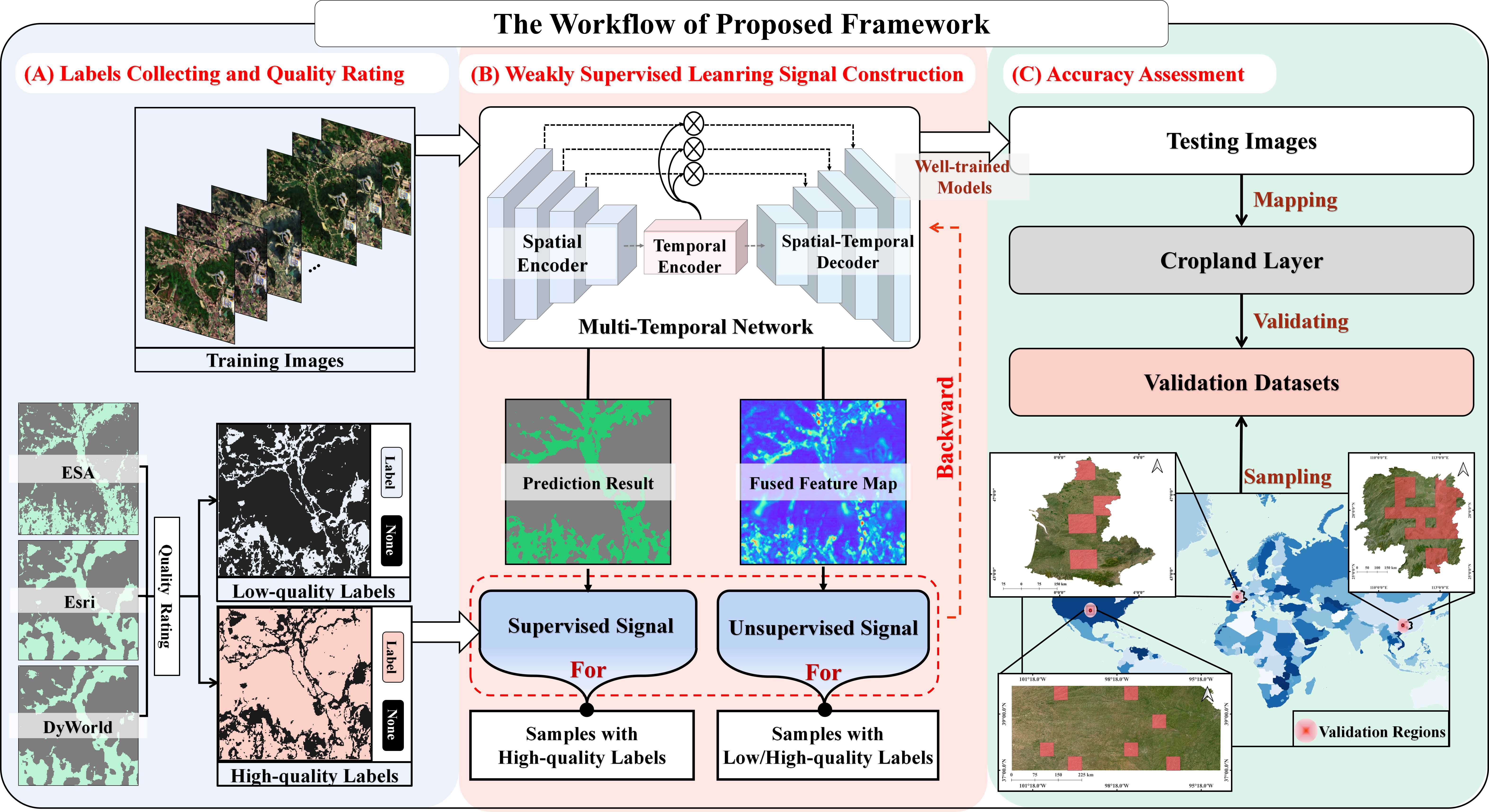}
	  \caption{The workflow of the proposed framework. (a) Labels collecting and quality rating, (b) Weakly supervised learning signal construction, (c) Accuracy assessment.}\label{fig3}
\end{figure}

The proposed framework aims to effectively use the prior information from existing GLC products for large-scale cropland mapping. By using this prior information, the model can learn the cropland phenology features from SITS without manual labeling. As shown in \cref{fig3}, the framework consists of three parts. \textbf{(1) Labels collecting and quality rating:} We collect labels from GLC products, and evaluate their quality. These labels are then categorized into high-quality and low-quality parts; \textbf{(2) Construction of weakly supervised learning signal:} we construct the supervised part of the learning signal using high-quality labels, and encode the image intrinsic feature distribution to construct the unsupervised part of the learning signal. By constructing the unsupervised part, we not only incorporate the abundant information contained in the samples with low-quality labels into the model learning process, but also prompt the model to assess and question the reliability of the high-quality labels. Additionally, we extend the weakly supervised signal in the temporal dimension to sufficiently extract the phenology features of cropland. \textbf{(3) Accuracy assessment:} we utilize the well-trained models for large-scale cropland mapping, and establish validation datasets to evaluate their performance.

\subsection{Labels collecting and quality rating}\label{Section 3.1}

\medskip

The cropland layers from GLC products are cross-referenced and quality-rated. Consistent parts are considered high-quality labels, and divergent parts are considered low-quality labels. The quality of a given spatial position $(i,j)$ is determined by whether the results of GLC products are the same. Finally, we obtained the high-quality sample mask ${M}_{high}(i,j)$ and high-quality labels $\mathcal{Y}(i,j)$:

\begin{equation}
M_{{high}}(i, j) =
\begin{cases}
    1, & \mathrm{if} \ P_1(i, j) = P_2(i, j) = \ldots = P_m(i, j) \\
    0, & \mathrm{otherwise}
\end{cases}\label{eq1}
\end{equation}

\begin{equation}
    \mathcal{Y}(i,j) = M_{{high}}(i, j)\ast\frac{1}{M}  {\textstyle \sum_{m=0}^{M}(P_m(i,j))}
\label{eq2}
\end{equation}
where $P_m(i,j)$ is the cropland labels of the m-th product at position $(i,j)$, and $M$ represents the total number of products

\subsection{The construction of weakly supervised learning signal}\label{Section 3.2}

\medskip

The weakly supervised learning signal consists of the supervised part and the unsupervised part. The supervised part is constructed based on high-quality labels, and it can effectively guide the model to learn the cropland features without manual labeling. The unsupervised part is constructed by encoding the visual similarity and spatial aggregation of the same land cover based on the feature space extracted from images. It serves as the regularization term to avoid over-fitting the remaining errors in high-quality labels. In addition, the unsupervised part integrates the abundant information contained in the low-quality labeled samples to enhance the diversity of the model’s feature space in the optimization process. 
\par

Given an image sequence $X$ of size $T\times C\times H\times W$, we use it as the input of the multi-temporal network to obtain the prediction result $P$, which is used to construct the supervised part of the learning signal. The intermediate feature maps extracted by the multi-temporal network are fused to form the feature space $Z$, which is used to construct the unsupervised part of the learning signal. Both the supervised and unsupervised parts are employed in samples with high-quality labels, but only the unsupervised part is employed in samples with low-quality labels. The specific construction process is as follows:
\par
\begin{figure}[htp]
	\centering
		\includegraphics[width=0.8\linewidth]{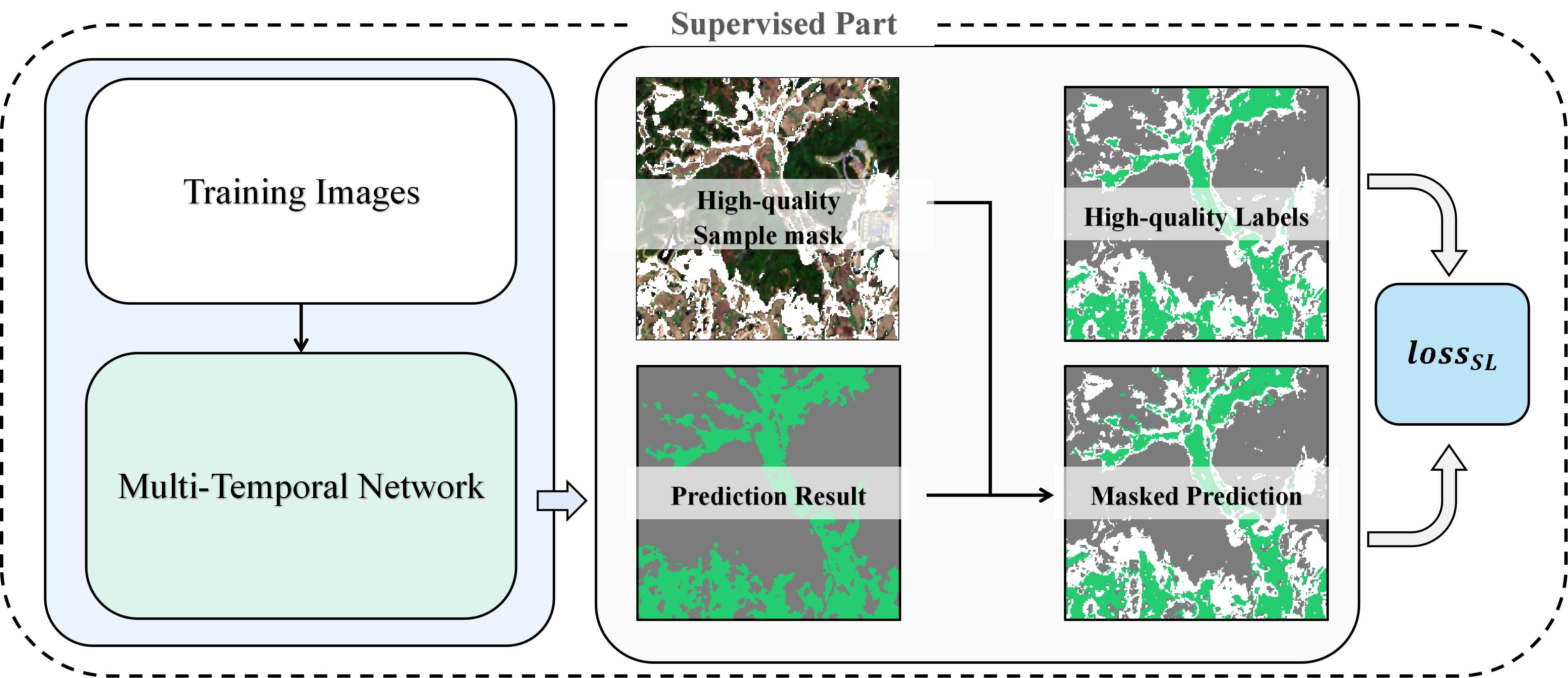}
	  \caption{ The flowchart to construct the supervised part of the learning signal.}\label{fig4}
\end{figure}

We use the high-quality sample mask and prediction result to generate the masked prediction, which is combined with the high-quality labels to construct the supervised part of the learning signal (\cref{fig4}). Given a pixel $(i,j)$, with $0\ \le\ i\ <\ H\ $ and $0\ \le\ j\ <\ W$, the masked prediction $M_{high}(i,j)\ast\mathcal{P}(i,j)$ and the high-quality $\mathcal{Y}(i,j)$ are used to calculate the supervised cross-entropy loss $Loss_{SL}$ as follows:

\begin{equation}
    {Loss}_{SL}=\ -\sum_{i\ =\ 0}^{H}\sum_{j\ =\ 0}^{W}{(\mathcal{Y}(i,j)\ast\log{(M_{high}(i,j)\ast\mathcal{P}(i,j))})}
\end{equation}

\begin{figure}[htp]
	\centering
		\includegraphics[width=0.95\linewidth]{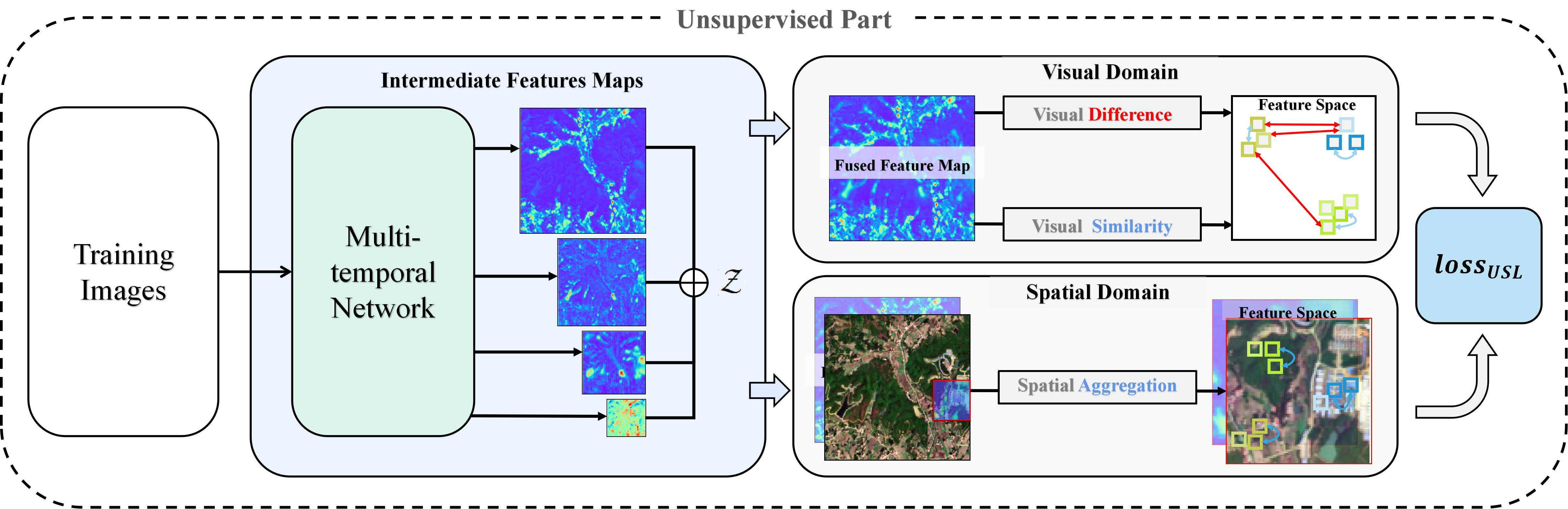}
	  \caption{ The flowchart to construct the unsupervised part of the learning signal.}\label{fig5}
\end{figure}

Due to the quality limitation of the products, the high-quality labels may contain some errors. To avoid over-fitting these errors, inspired by \citet{hua2021semantic} and \citet{sabokrouSelfSupervisedRepresentationLearning2019a}, we construct the unsupervised part of the learning signal as the regularization term to constrain the model. The construction is based on two assumptions: (1) In the visual domain, two visually similar samples have a higher probability of belonging to the same semantical concept \citep{sabokrouSelfSupervisedRepresentationLearning2019a}, which means these samples are adjacent in the model’s feature space. (2) In the spatial domain, the land covers are continuous and aggregated \citep{jiangGeospatialAnalysisRequires2015}, which means the adjacent samples with the most similarities should belong to the same category. The regularization term allows the model to enhance the stability of the feature space during the optimization process, which can alleviate cognitive bias caused by the over-fitting of the remaining errors.\par
To enhance the model's generalization ability in large-scale cropland mapping, we also employ the unsupervised part of the learning signal in the low-quality samples. Thus, the information from these samples can be involved in the model optimization process, balancing the intra-class feature distribution of the training samples and improving the diversity of the feature space.\par

In \cref{fig5}, the feature space $\mathcal{Z}$ is represented by the fused intermediate feature maps obtained from the multi-temporal network. Given the sample $x_n$, which represents the nth pixel in an image mapped to the high-dimensional feature space $\mathcal{Z}$, where $N = W \times H$ and $n \in \mathbb{N}$. In the visual domain, the sample $x_n^s$ and $x_{n}^d$ are identified as the samples with the highest similarity and difference, respectively, to $x_n$. These are determined by searching within the same image using the Sorensen-Dice index. We then encourage the feature distance between $x_n$ and $x_n^s/x_{n\ }^d $ to be as small/large as possible during model optimization In the spatial domain, given $x_n^{sn}$ as the sample with the highest similarity to $x_n$ among its eight-neighborhood samples, we encourage the model to minimize their feature distance. Ultimately, the constraints from both the visual and spatial domains are integrated to yield the unsupervised loss ${Loss}_{USL}$:

\begin{equation}
        {Loss}_{USL} = \alpha \sum_{n=0}^{N} D_{KL}[{\mathcal{Z}}(x_n),{\mathcal{\mathcal{Z}}}(x^s_n)]  - \beta \sum_{n=0}^{N} D_{KL}[{\mathcal{Z}}(x_n),{\mathcal{\mathcal{Z}}}(x^d_n)]  +  \gamma \sum_{n=0}^{N} D_{KL}[{\mathcal{Z}}(x_n),{\mathcal{Z}}(x^{sn}_n)]
\end{equation}
where $\alpha$, $\beta$, and $\gamma$ are the priori parameters to measure the importance of different terms. $D_{KL}$ represents the Kullback-Leibler Divergence. At last, the entire model is optimized by minimizing the weakly supervised loss $Loss_{WS}$ that combines the aforementioned supervised and unsupervised parts:

\begin{equation}
    {Loss}_{WS}\ = {Loss}_{SL}+\ {Loss}_{USL}
\end{equation}

\subsection{The structure of multi-temporal network}\label{Section 3.3}

\medskip

In this paper, considering the temporal features of cropland, we use the multi-temporal network for phenological feature extraction to enhance the separability of cropland in the feature space. Note that, the multi-temporal network is changeable, allowing for different network architectures to be employed or substituted as needed. We select U-Net with Temporal Attention Encoder (U-TAE) \citep{faregarnotPanopticSegmentationSatellite2021b} to extract multi-scale spatio-temporal features, which enhances the robustness of anomalies and the capture ability of long-term dependencies in temporal features. The U-TAE contains three parts: spatial encoder, temporal encoder, and spatial-temporal decoder.
\par

In the spatial encoder, each image in the temporal sequence is embedded by a shared multi-level convolutional spatial encoder $\mathbb{E}^l$. In Equation (\ref{eq3}), image sequence $X$ of size $T\times C\times H\times W$ is used as input, The multi-scale spatial feature sequence $e^l$ of the temporal image is obtained by continuously down-sampling using sliding convolution at each layer.

\begin{equation}
    {e}^l=\begin{cases}
X, & l=0
\\
\mathbb{E}^l(e^{l-1})^{T}_{t=0} , &  \mathrm{for} \ l \in [1,L]
\end{cases}\label{eq3}
\end{equation}
where l is the number of layers and the size of $e^l$ is $T\times {C^{l}}\times {H^{l}}\times {W^{l}}$.
\par

In the temporal encoder, Lightweight-Temporal Attention Encoder (L-TAE) \citep{garnotLightweightTemporalSelfattention2020b} is used for processing the temporal dimension, which can help the model to capture long-term dependencies and adapt to dynamic temporal features of the cropland. In this process, the model obtains attention $a^l$ of the image sequence in the temporal dimension based on the lowest scale spatial feature, and resizes it to ${H^l}\times{W^l}$ for compressing the sequence of spatial features at different scales. Eventually, the feature sequences $\mathcal{F}^{l}$ with size ${C^{l}}\times {H^{l}}\times {W^{l}}$ are obtained from multi-scale spatial features $e^l$ and the temporal attention ${a^l}$ :

\begin{equation}
    {a^l}=\begin{cases}
\text{LTAE}(e^l), & \mathrm{if} \ l=L
\\
\text{resize}[\text{LTAE}(e^l)]^{L-1}_{l=0}, & \mathrm{for} \ l\in [0,L-1]
\end{cases}
\end{equation}


\begin{equation}
    \mathcal{F}^{l}= {\textstyle \sum_{t=0}^{T}}Conv^{l}_{1\times 1}[a^l_t\odot e^l_t]^L_{l=0}, \
\mathrm{for} \ l\in [0,L]
\end{equation}
where $Conv^{l}_{1\times 1}$ is a shared $1\times1$ convolution layer of width $C^l$ and $\odot$ is the term-wise multiplication with channel broadcasting.\par



In the Spatial-Temporal decoding part, a multi-level convolutional decoder $\mathbb{D}^l$ is used to generate single spatial-temporal feature maps on different scales. In detail, the feature map $\mathcal{D}^l$ connects with compressed features $\mathcal{F}^l$ channel-wise, and continuously up-sampled by transposed convolution $\mathbb{D}^l_{up}$ to get the multi-scale spatial-temporal feature maps:


\begin{equation}
\mathcal{D}^l=
\begin{cases}
\mathcal{F}^L \ ,&\mathrm{if} \ l=L
\\
\mathbb{D}^l(\mathbb{D}^l_{up}[(\mathcal{D}^{l-1})],\mathcal{F}^l)\ , & \mathrm{for} \ l \in [0,L-1]
\end{cases}
\end{equation}

\begin{equation}
    \mathcal{P} = softmax(Conv(\mathcal{D}^L))
\end{equation}
where $\left[\bullet\right]$ is the channel-wise concatenation. The last feature map $\mathcal{D}^L$ with the same size as the original image $(H\times W)$ is processed by the convolution layer $Conv(\bullet)$ and the activation function $softmax(\bullet)$ to get the predictions $\mathcal{P}$.
\par
Ultimately, the predictions $\mathcal{P}$ and multi-scale spatial-temporal feature maps 
 $\mathcal{D}^l$  are used as  $\mathcal{Z}$  for constructing the supervised and unsupervised part of the learning signal, respectively.

\subsection{Mapping and accuracy assessment}\label{Section 3.4}

\medskip
Given that the proposed method does not rely on manual annotations for model training, we are not constrained by the typical limitations of training and validation set independence. Although we refer to the information from GLC products as labels, they more closely resemble pseudo-labels. The overlap between the training and validation sets will not affect the accuracy assessment process. Consequently, we trained the models with each study area, and employed the well-trained models in three study areas (Hunan Province in China, Southeast France, and Kansas State in the USA) to map the cropland  (\cref{fig3}). Meanwhile, we carefully sampled the representative regions of each study area for manual labeling, and constructed the validation set, which is included in the region of the training set. To ensure a comprehensive and objective assessment, the selection process for the validation area was guided by topography to balance different cropland scenes (plains, hills, and mountainous). Details regarding the selection and labeling of the validation area are described in \ref{Section 2.2}.

To evaluate the accuracy of our cropland mapping results, we employed a set of assessment metrics \citep{olofsson2014good,li2023development}, including Overall Accuracy (OA), Producer’s Accuracy (PA), User’s Accuracy (UA), Mean Intersection over Union (mIoU) \citep{deng2021cnns}, the macro-average of F1-scores across all (Avg. F1-score), and the F1-score of cropland (Crop. F1-score ) and Non-cropland (Non. crop. F1-score) \citep{zhong2019deep}. The F1-score acts as a harmonic mean that integrates PA and UA to measure the model’s precision and recall capabilities effectively.

\section{Experiment}\label{Section 4}

\subsection{Experiment setting}\label{Section 4.1}

\medskip

Experiments used the datasets collected from Hunan province in China, Southeast France, and Kansas state in the USA. We cropped all the images into the size of 256*256 pixels. For the training dataset, we produced 32,318 patches in the Hunan study area, 29,893 in Southeast France, and 32,515 in Kansas. For the validation dataset, we obtained 15,484 patches within the Hunan study area, 6,152 in Southeast France, and 7,697 in Kansas. In the process of cropland mapping, we segmented all the images into multiple patches with a sliding length of 128 pixels, and utilized probabilistic prediction results to integrate the final result.


All models were trained using PyTorch on the Ubuntu 16.04 operation system with an NVIDIA GTX3080 GPU (11-GB memory). Each model was trained using the Adam optimizer, with a batch size of 8 and 100 epochs. The learning rate was initially set to $ {1\times10}^{-3}$ and decrease it to ${1\times10}^{-4}$ for the last 50 epochs.

\begin{figure}[ht]
	\centering
		\includegraphics[width=1\linewidth]{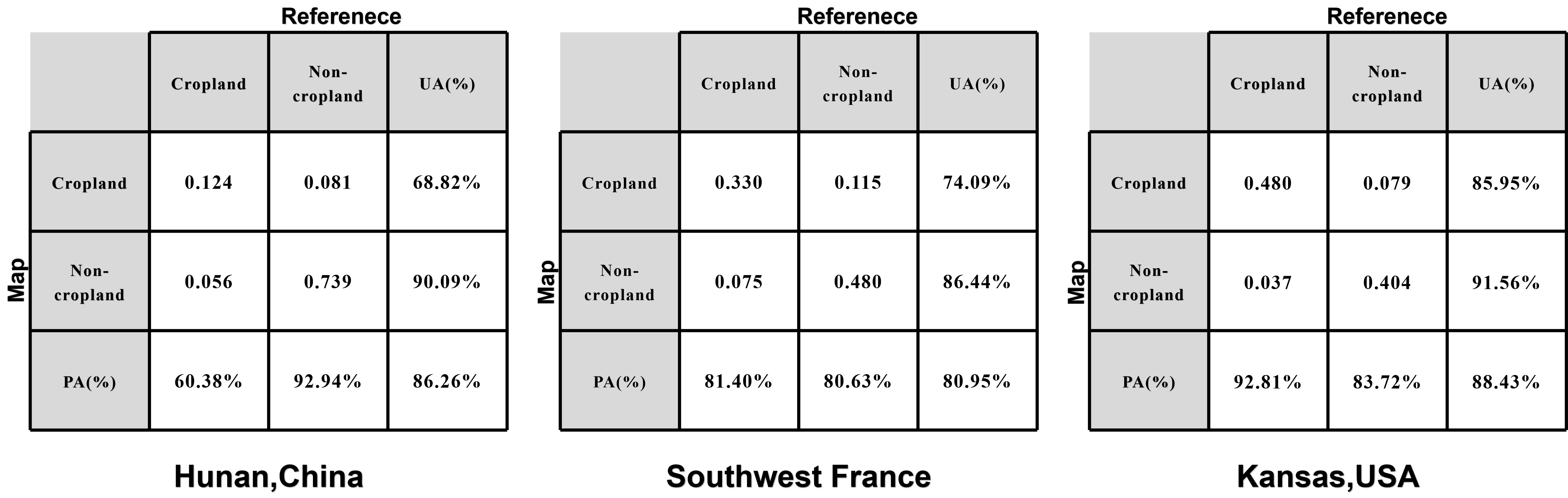}
	  \caption{ Confusion matrices represented by area ratios for Hunan, Southwest France, and Kansas study areas}\label{fig6}
\end{figure}

\subsection{Cropland mapping results}\label{Section 4.2}

\medskip

\par
\begin{figure}[htp]
	\centering
		\includegraphics[width=0.75\linewidth]{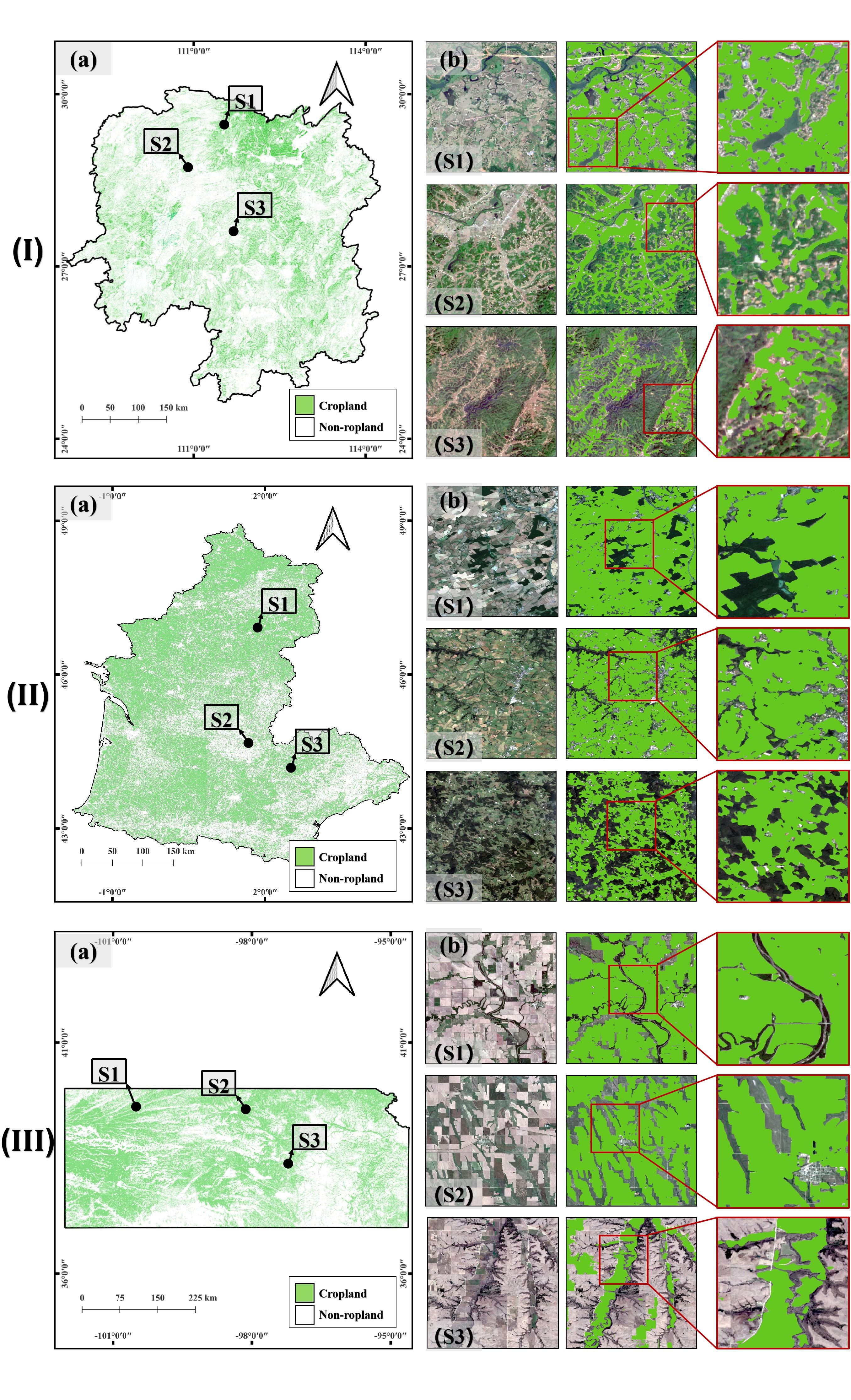}
	  \caption{ (a) Overall cropland mapping results for Hunan, Southwest France, and Kansas in 2020, (b) the image and classification results for plain cropland (S1), hill cropland (S2), and mountain cropland (S3), respectively.}\label{fig7}
\end{figure}

To demonstrate the effectiveness of our framework, we present the confusion matrix and all accuracy evaluation metrics for our cropland mapping results across the three study regions in \cref{fig6} and \cref{tbl2}. In the study areas of Hunan, Southwest France, and Kansas, we attained an Avg.F1-score of 77.91\%, 80.50\%, and 88.36\%, respectively. It is observable that compared to other study areas, the accuracy of the Hunan is relatively low. This is attributed to the fact that the farming pattern in Hunan is predominantly smallholder, which is distinct from the large-scale agricultural operations common in Europe and America. Meanwhile, the prevalence of hilly and mountainous terrain in the Hunan region leads to smaller average field sizes, a more fragmented spatial distribution, and a greater diversity of cropland types. These factors collectively render the classification of croplands in this area more challenging. We also visualized the cropland mapping result in \cref{fig7}(a), and selected typical samples of three terrain types (plains, hills, and mountains) in each study area to demonstrate the details \cref{fig7}(b). Specifically, we determined the terrain types by analyzing the slope of each pixel using the Digital Elevation Model (DEM) from STRM. 
All samples were categorized into plains (0° to 2°), hills (2° to 6°), and mountains (greater than 6°). As shown in \cref{fig7}, our framework successfully extracts plain croplands, which tend to have relatively large average field size, exhibiting a clear distinction between built-up areas and rivers. Hills croplands and mountain croplands exhibit relatively small average field sizes, which typically exhibit a fragmented distribution mixed with other types of vegetation cover, and our framework is also capable of accurately identifying their boundaries. 


\par

\begin{figure}[htp]
	\centering
		\includegraphics[width=0.75\linewidth]{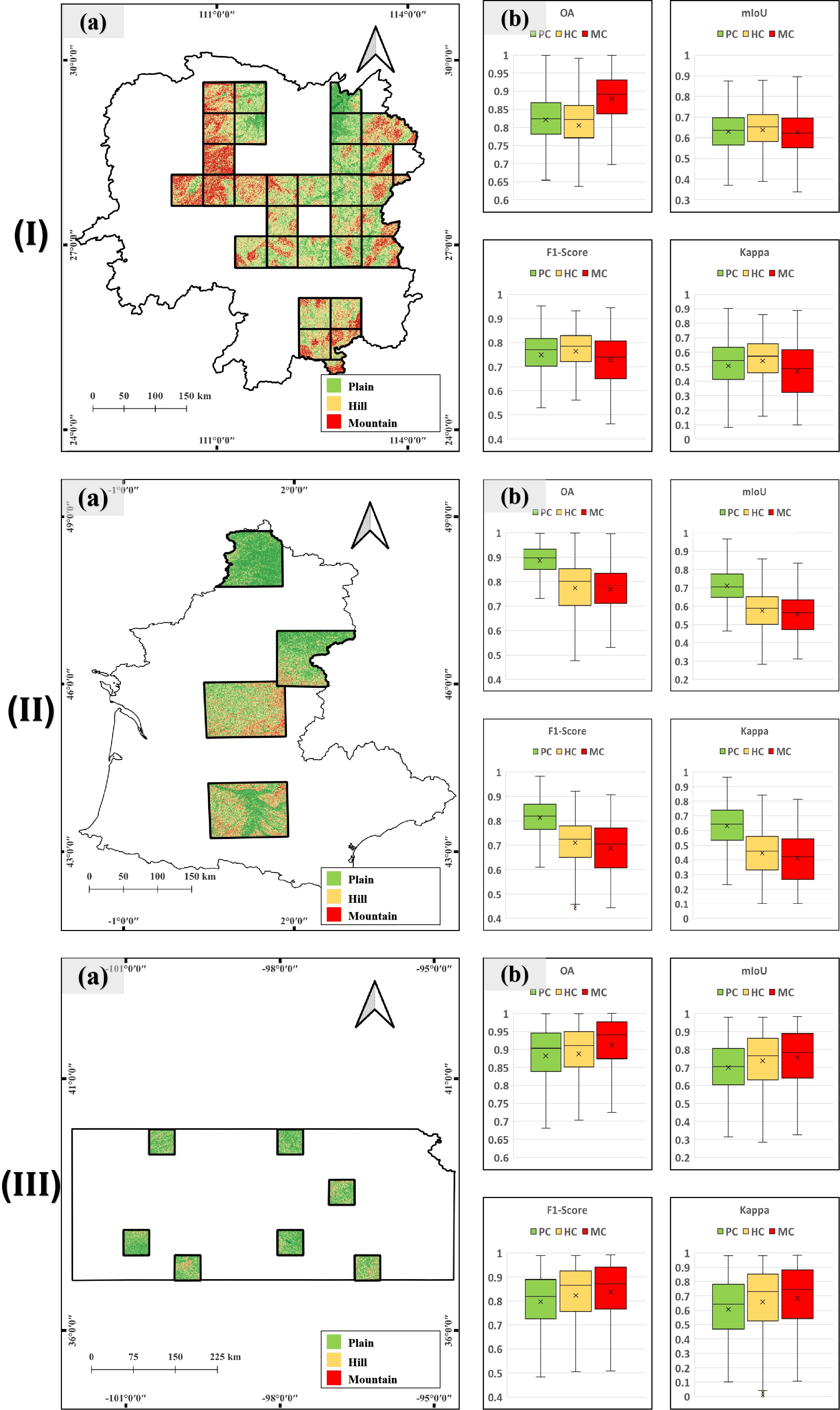}
	  \caption{ (a) Classification map of plain croplands (PC), hill croplands (HC), and mountain croplands (MC) in the validation regions of the three study areas. (b) the boxplots of main evaluation metrics in different types of croplands. “×” denotes the location of the average value.}\label{fig8}
\end{figure}

Furthermore, we analyzed the results using key evaluation metrics for the plain croplands (PC), hill croplands (HC), and mountain croplands (MC) in each study area (\cref{fig8}). In Hunan study areas, the Avg.F1-score is 74.85\%, 76.37\%, and 72.84\% for plain, hill, and mountain croplands, respectively. The variability is lower in the hill cropland than in the other two. In the Southwest France study areas, the Avg.F1-score is 81.41\%, 71.00\%, and 68.76\% for plain, hill, and mountain croplands, respectively. The fluctuations are smaller in plain croplands than in hill and mountain croplands. In the Kansas study areas, the Avg.F1-score accuracy is 79.75\% for plain croplands, 82.29\% for hill croplands, and 83.17\% for mountainous croplands. This counterintuitive phenomenon can be attributed to the incorporation of temporal information. Under the SITS observations, in hilly and mountainous areas, other types of vegetation exhibit more distinct phenological differences from croplands than in flat regions. Conversely, in the plains, some croplands demonstrate similar phenological patterns to other vegetation covers, such as shrubs or grasses.

\begin{table}[h]
\caption{The accuracy evaluation of our cropland mapping results across three study areas}\label{tbl2}
{
\scriptsize
\renewcommand{\arraystretch}{1.6}
\begin{tabular*}{\tblwidth}{@{}LLLLLLLL@{}}
\hline
\multirow{2}{*}{Study Areas} & \multirow{2}{*}{OA(\%)} & \multirow{2}{*}{mIoU(\%)} & \multirow{2}{*}{Avg.F1-score(\%)} & \multicolumn{2}{l}{Non-cropland} & \multicolumn{2}{l}{Cropland} \\ \cline{5-8} 
                            &                         &                           &                               & PA(\%)        & UA(\%)       & PA(\%)          & UA(\%)         \\ \hline
Hunan,China                 & 86.26\%                 & 65.86\%                   & 77.91\%                       & 90.09\%       & 92.94\%      & 68.82\%         & 60.38\%        \\
Southwest France            & 80.95\%                 & 67.47\%                   & 80.50\%                       & 81.40\%       & 74.09\%      & 80.63\%         & 86.44\%        \\
Kansas,USA                  & 88.43\%                 & 79.15\%                   & 88.36\%                       & 92.81\%       & 85.95\%      & 83.72\%         & 91.56\%        \\ \hline
\end{tabular*}}

\end{table}

\begin{table}[h]
\caption{Cropland mapping accuracy of our framework and the cropland layers across the three GLC products.}\label{tbl3}
{
\scriptsize
\renewcommand{\arraystretch}{1.3}
\begin{tabular*}{\tblwidth}{@{}LLLLLLL@{}}
\toprule

Study Areas                           & Products & OA(\%)           & mIoU(\%)         & Avg.F1-score(\%) & Crop.F1-score(\%) & Non-crop.F1-score(\%) \\ \midrule 
\multirow{4}{*}{Hunan Province,China} & ESA      & 83.46\%          & 60.85\%          & 73.60\%              & 57.47\%           & 89.73\%               \\
                                      & Esri     & 80.26\%          & 55.37\%          & 68.44\%              & 49.13\%           & 87.75\%               \\
                                      & DyWorld  & 85.07\%          & 59.95\%          & 72.15\%              & 53.17\%           & 91.12\%               \\
                                      & Ours     & \textbf{86.26\%} & \textbf{65.86\%} & \textbf{77.91\%}     & \textbf{64.32\%}  & \textbf{91.49\%}      \\ \midrule 
\multirow{4}{*}{Southwest France}     & ESA      & 71.50\%          & 55.61\%          & 71.47\%              & 70.56\%           & 72.37\%               \\
                                      & Esri     & 80.91\%          & 66.99\%          & 80.10\%              & 83.13\%           & 77.06\%               \\
                                      & DyWorld  & 73.68\%          & 58.32\%          & 73.67\%              & 74.21\%           & 73.12\%               \\
                                      & Ours     & \textbf{80.95\%} & \textbf{67.47\%} & \textbf{80.50\%}     & \textbf{83.44\%}  & \textbf{77.57\%}      \\ \midrule 
\multirow{4}{*}{Kansas,USA}           & ESA      & 87.32\%          & 77.24\%          & 87.13\%              & 85.60\%           & 88.67\%               \\
                                      & Esri     & 80.86\%          & 67.86\%          & 80.85\%              & 81.18\%           & 80.52\%               \\
                                      & DyWorld  & 78.96\%          & 65.23\%          & 78.95\%              & 79.31\%           & 78.60\%               \\
                                      & Ours     & \textbf{88.43\%} & \textbf{79.15\%} & \textbf{88.36\%}     & \textbf{87.47\%}  & \textbf{89.25\%}      \\ \bottomrule
\end{tabular*}
}

\end{table}

\begin{figure}[htp]
	\centering
		\includegraphics[width=1\linewidth]{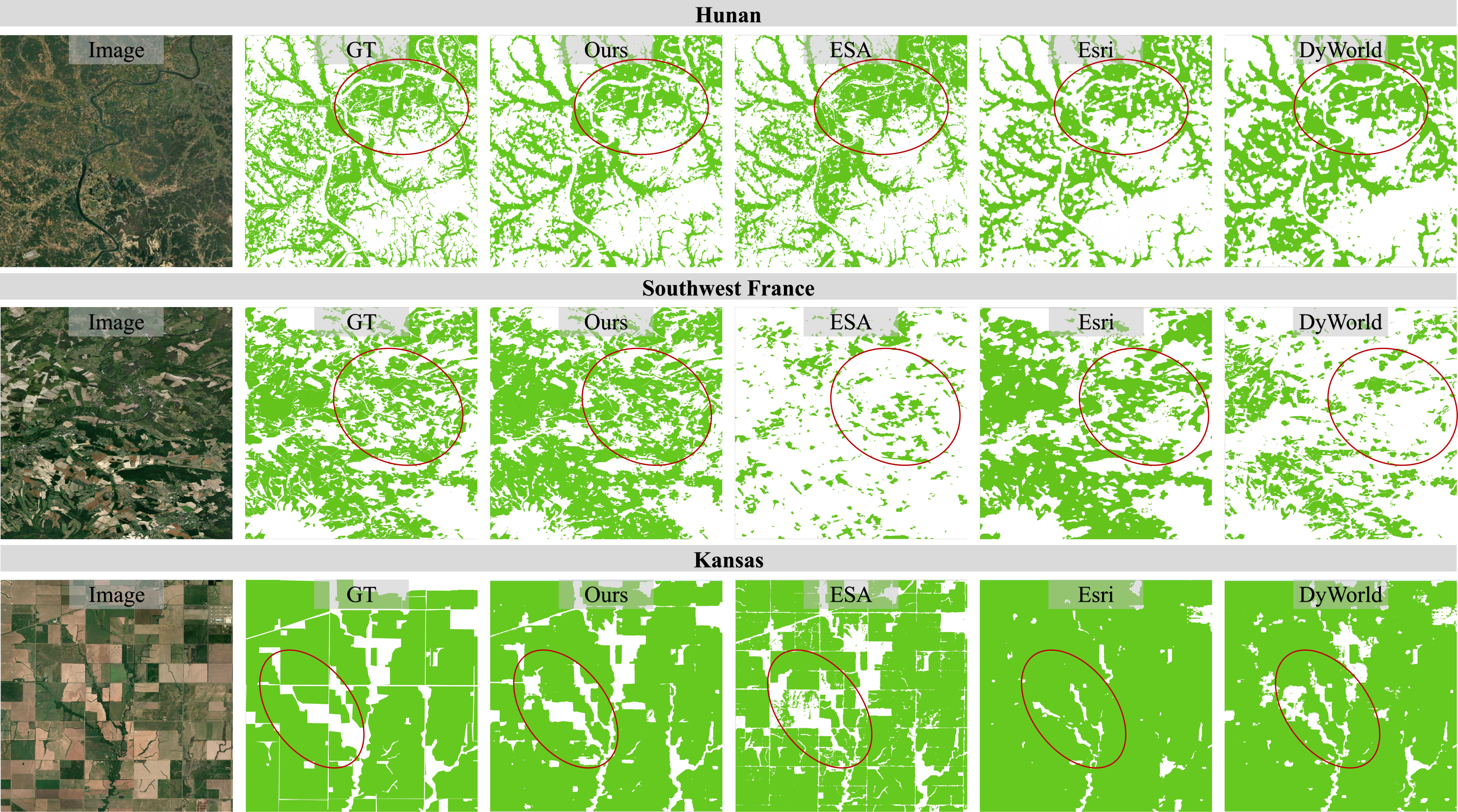}
	  \caption{Cropland mapping results of our method and the cropland layers of other GLC products.}\label{fig9}
\end{figure}

\subsection{Comparison with other GLC products}\label{Section 4.3}

\medskip
A great challenge for large-scale cropland mapping is to ensure the framework’s generalizability across diverse scenarios while maintaining low labeling costs. While our framework provides a promising solution to reduce labeling costs through the proposed weakly supervised learning signals, its generalizability in practical scenarios needs objective assessment. Therefore, we compared the proposed approach with three public GLC products (ESA, Esri, and Dyworld) by a series of comparative experiments across diverse agrosystems, including Hunan, Southwest France, and Kansas. In quantitative terms, the results in \cref{tbl3} demonstrate that the proposed framework outperformed the three products. Compared with the highest accuracy of GLC products, our framework achieved improvements in the Avg.F1-score by 5.84\%, 0.51\%, and 1.40\% in the corresponding study area, while achieving improvements in the Crop.F1-score by 11.91\%, 0.37\%, and 2.18\%. The qualitative comparison results are shown in \cref{fig9}, for the Hunan study area, our results are more comprehensive and provide better extraction of fragmented plain and hill croplands. For the Southwest France study area, our results exhibit greater detail. In the Kansas study area, our method is capable of extracting fields with unusual phenological attributes.

\begin{figure}[ht]
	\centering
		\includegraphics[width=1\linewidth]{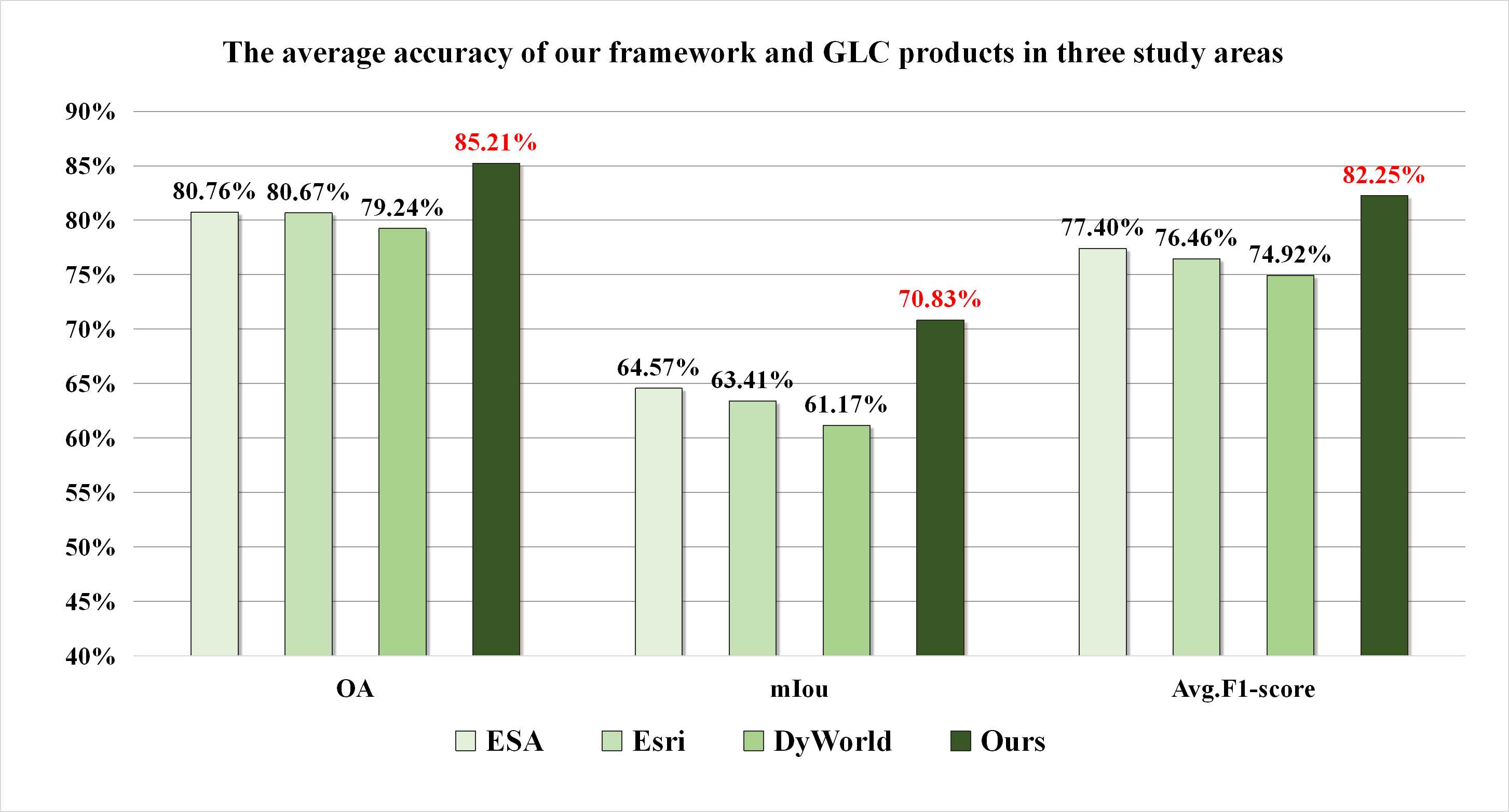}
	  \caption{The average accuracy of our framework and the cropland layers from ESA, Esri, and Dyworld in three study areas.}\label{fig10}
\end{figure}

Moreover, the stability of the model’s performance is crucial for large-scale cropland mapping in practical application scenarios. However, as shown in \cref{tbl3}, the accuracy of the three products varies greatly across these study areas. For instance, the ESA performs exceptionally well in Kansas, but it showed the poorest performance in Southwest France. To further evaluate the reliability, we calculated the average accuracy of our proposed framework and the three products in the three study areas (\cref{fig10}). Our framework demonstrates notable improvements in terms of OA, mIoU, and Avg.F1-score, surpassing the best-performing products by 5.52\%, 9.70\%, and 6.27\%, respectively. These results clearly indicated the superior reliability and stability of our proposed framework.

\begin{table}[h]
\caption{The classification accuracy of DeeplabV3+, Unet-3D, LSTM, U-TAE, RRE, WESUP-LCP and our method in the study areas of Hunan, Southwest France, and Kansas.}\label{tbl4}
{
\scriptsize 
\renewcommand{\arraystretch}{1.6}
\begin{tabular*}{\tblwidth}{@{}LLLLLLL@{}}
\toprule
Study Areas                            & Methods    & OA(\%)               & mIoU(\%)            & Avg.F1-score(\%)     & Crop.F1-score(\%)    & Non-crop.F1-score(\%) \\ \midrule 
\multirow{7}{*}{Hunan Province,China} & DeeplabV3+ & 85.56\%          & 61.78\%          & 73.98\%          & 56.63\%          & 91.34\%           \\
                                      & Unet-3D    & 85.57\%          & 62.44\%          & 74.68\%          & 58.07\%          & 91.29\%           \\
                                      & LSTM       & 86.02\%          & 63.15\%          & 75.28\%          & 58.99\%          & 91.58\%           \\
                                      & U-TAE      & 86.38\%          & 63.63\%          & 75.68\%          & 59.55\%          & \textbf{91.81\%}           \\
                                      & RRE        & 85.93\%          & 62.98\%          & 75.13\%          & 58.74\%          & 91.52\%  \\
                                      & WESUP-LCP  & 86.14\%          & 63.71\%          & 75.82\%          & 60.03\%          & 60.03\%           \\
                                      & Ours       & \textbf{86.26\%}          & \textbf{65.86\%}          & \textbf{77.91\%}          & \textbf{64.32\%} & 91.49\%           \\ \midrule 
\multirow{7}{*}{Southwest France}     & DeeplabV3+ & 75.84\%          & 60.97\%          & 75.73\%          & 77.39\%          & 74.06\%           \\
                                      & Unet-3D    & 78.34\%          & 64.23\%          & 78.19\%          & 80.02\%          & 76.36\%           \\
                                      & LSTM       & 78.32\%          & 64.14\%          & 78.12\%          & 80.19\%          & 76.05\%           \\
                                      & U-TAE      & 77.70\%          & 63.37\%          & 77.55\%          & 79.36\%          & 75.75\%           \\
                                      & RRE        & 77.30\%          & 62.83\%          & 77.14\%          & 79.02\%          & 75.26\%           \\
                                      & WESUP-LCP  & 76.96\%          & 62.40\%          & 76.82\%          & 78.62\%          & 75.03\%           \\
                                      & Ours       & \textbf{80.95\%} & \textbf{67.42\%} & \textbf{80.47\%} & \textbf{83.44\%} & \textbf{77.57\%}  \\ \midrule 
\multirow{7}{*}{Kansas State, USA}    & DeeplabV3+ & 86.82\%          & 76.63\%          & 86.76\%          & 85.85\%  & 87.66\%                   \\
                                      & Unet-3D    & 88.14\%          & 78.70\%          & 88.07\%          & 87.18\%  & 88.96\%                     \\
                                      & LSTM       & 88.05\%          & 78.56\%          & 87.98\%          & 87.09\%  & 88.87\%                  \\
                                      & U-TAE      & 87.40\%          & 77.56\%          & 87.36\%          & 86.64\%  & 88.07\%                   \\
                                      & RRE        & 87.41\%          & 77.53\%          & 87.33\%          & 86.33\%  & 88.34\%                   \\
                                      & WESUP-LCP  & 87.35\%          & 77.33\%          & 87.20\%          & 85.81\%  & 88.59\%                     \\
                                      & Ours       & \textbf{88.43\%} & \textbf{79.15\%} & \textbf{88.36\%} & \textbf{87.47\%} & \textbf{89.25\%}   \\ \midrule 
\end{tabular*}
}
\end{table}

\begin{figure}
	\centering
		\includegraphics[width=1\linewidth]{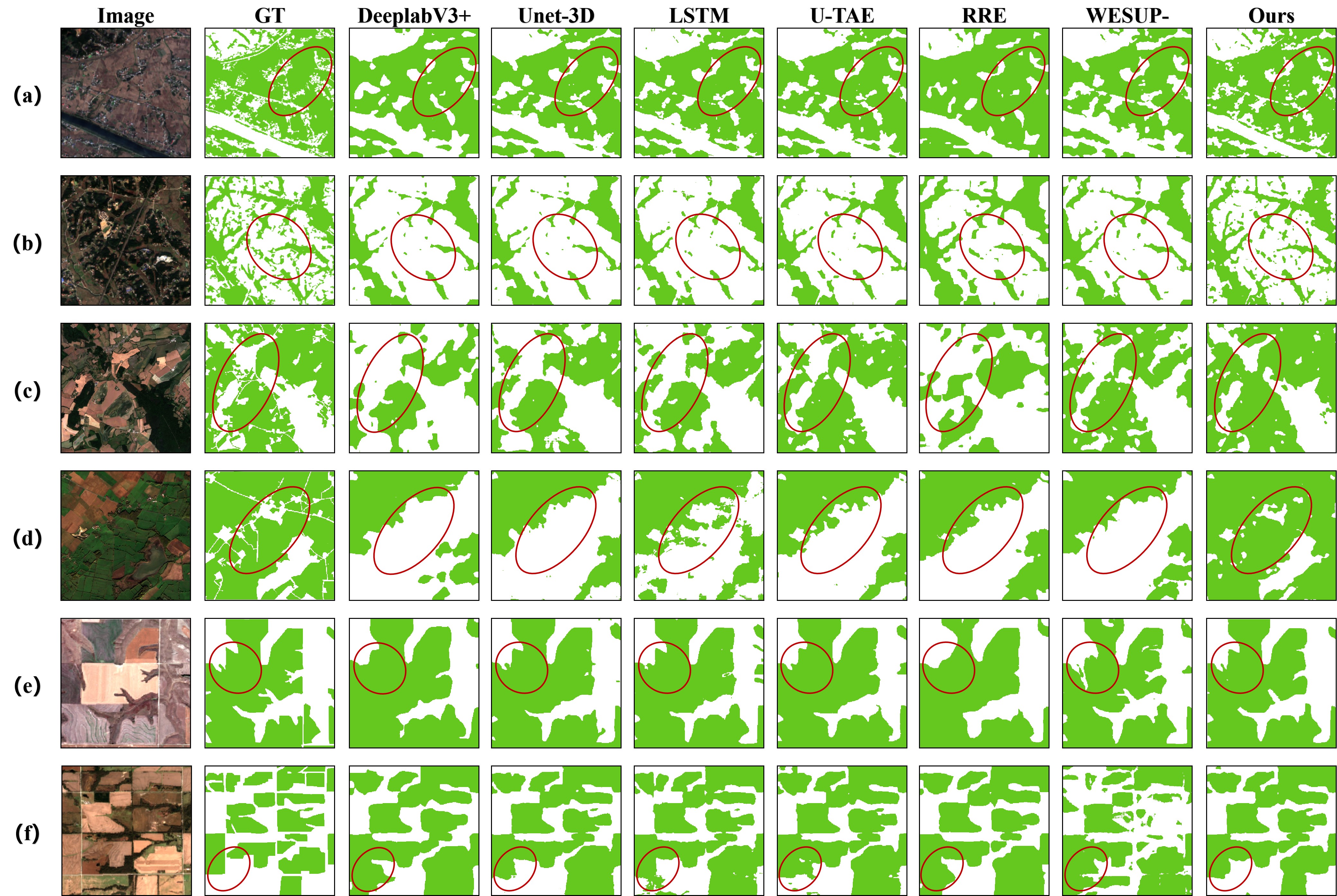}
	  \caption{The classification results of DeeplabV3+, Unet-3D, LSTM, U-TAE, RRE, WESUP-LCP, and our method in the three study areas. (a) and (b) for Hunan, (c) and (d) from Southwest France, (e) and (f) for Kansas.}\label{fig11}
\end{figure}

\subsection{Comparison with other methods}\label{Section 4.4}

\medskip

To demonstrate the superiority of the proposed method, we compared it with the following three types of methods based on Automatic Training Sample Generation (ATSG) :
\par
\textbf{Discard methods}: We used classical single- and multi-temporal networks, including Deeplabv3+ \citep{chenEncoderDecoderAtrousSeparable2018a}, Unet-3D \citep{Rustowicz2019SemanticSO} and LSTM \citep{shi2015convolutional}, and U-TAE as classifiers, relying solely on supervised signals derived from high-quality labels to guide their learning process. We took the year-round composite images as the base data for single-temporal network, and used the dense SITS for multi-temporal networks. 
\par
\textbf{Re-correct method}: We utilized the strategy from the RRE  framework \citep{zhangNovelKnowledgeDrivenAutomated2023a} to construct the comparison method, which is an automated solution for extracting high-resolution cropland through cross-scale sample transfer. First, we trained a label corrector using only high-quality labels and multi-temporal images to obtain the labels of low-quality samples. Then the corrected labels were used to generate supervised signals to guide the model learning process. 
\par
\textbf{Re-correct method with weakly supervised learning}: We chose the WESUP-LCP \citep{chenNovelWeaklySupervised2023b} as the comparison method, which is a weakly supervised semantic segmentation network for product resolution enhancement based on the re-correct method. This method is also available for our task. Specifically, we used the same filtering strategy as WESUP-LCP to identify high-quality sample points, which were then extended to pixels with low-quality labels by the super-pixel method. Then, we took the super-pixel labels to construct the supervised learning signals and used the deep dynamic label propagation mechanism to generate pseudo-labels for constructing weakly supervised signals.

\par
As the results in \cref{tbl4} show, our method achieved the best accuracy across most assessment metrics in all three study areas. Our method outperformed other approaches by achieving improvements of 3.38\%, 5.05\%, and 0.58\% in mIoU, and 7.15\%, 4.05\%, and 0.33\% in crop.F1-score for the Hunan, Southwest France, and Kansas study areas, respectively. The qualitative visual results in \cref{fig11} also demonstrate that our method outperformed the others in terms of completeness and the ability to capture detailed information across all three study areas. 
\par
In the Hunan study areas, our method (\cref{fig11}. (a) and \cref{fig11}. (b)) outperformed both the discard and re-correct methods in mapping fragmented croplands with small average field sizes. The discard method limited the model’s ability to learn the abundance of information from the region with low-quality labels, resulting in overfitting to cropland features with large average field sizes and a failure to recognize fragmented croplands with small average field sizes. The re-correct methods introduced errors into high-quality labels, which were then amplified during label propagation, leading the model to misclassify other land covers as cropland. The WESUP-LCP was designed for increasing label resolution and it was based on the re-correct method. Although it grasped more details than the other methods, it was still unable to accurately identify fragmented croplands with small average field sizes. 
\par
In the Southwest France study area, our method (\cref{fig11}. (c) and \cref{fig11}. (d)) extracted croplands with significantly different features. Other methods only identified the croplands that were not planted in a specific period (showed yellow soil), but omitted the planted croplands in the same period. This is because different GLC products present great inconsistency in the cropland regions with unusual phenological attributes, which makes it difficult to obtain high-quality labels for these samples. For the discard methods, the absence of labels caused the model to solely focus on typical cropland features, but lose the ability to recognize diverse cropland features. In addition, due to a lack of reliable references, the re-correct methods may generate a lot of incorrect labels for cropland areas with unusual phenological attributes. This consequently leads to the model overfitting inaccurate information. 
\par
In the Kansas study areas (\cref{fig11}. (e) and \cref{fig11}. (f)), the croplands have large average field sizes and share similar features, facilitating the generation of a large number of high-quality labels. Therefore, both the discard and re-correct methods performed well in these areas. Nevertheless, there is still the risk of misclassifying other vegetation cover as cropland. In contrast, the proposed method was capable of accurately detecting the boundary between farmland and other vegetation covers, such as lawns and shrubs. 
\par
Additionally, as shown in \cref{tbl4}, the networks that incorporate multi-temporal information exhibit significant advantages over the single-temporal network, indicating that using time-series information enables the model to enhance its ability to distinguish croplands from other land covers. We further discussed the necessity of using temporal information in \cref{Section 5.2}.

\section{Discussion}\label{Section 5}
In this section, we conducted ablation experiments across all three study areas to comprehensively analyze the input setting, and further discussed the limitations and effects of using temporal information under our framework. We discussed four questions: the impacts of using different GLC product combinations as inputs, the temporal generalizability of our framework, the benefit of expanding the framework in the temporal dimension, and the robustness of the proposed framework in real-world scenarios.
\subsection{Analysis of Employing Varied GLC Product Combinations as Inputs}\label{Section 5.1}

\medskip

\begin{table}
\caption{ The number/accuracy of high-quality labels and the accuracy of the final prediction results obtained by using different combinations of GLC products as inputs.}\label{tbl5}
{
\scriptsize 
\renewcommand{\arraystretch}{1.6}
\begin{tabular*}{\tblwidth}{@{}LCCCLLL@{}}
\hline
\multirow{3}{*}{Study Areas}          & \multicolumn{5}{c}{Inputs}                                                                                    & \multirow{3}{*}{Prediction Avg.F1-score} \\ \cline{2-6}
                                      & \multicolumn{3}{c}{Combination of Products} & \multirow{2}{*}{Label Avg.F1-score} & \multirow{2}{*}{Label Ratio} &                                   \\ \cline{2-4}
                                      & DyWorld         & ESA         & Esri        &                                 &                              &                                   \\ \hline
\multirow{4}{*}{Hunan Province,China} &                 & $\scriptstyle\surd$           & $\scriptstyle\surd$           & 78.87\%                         & 67.99\%                      & 74.69\%                           \\
                                      & $\scriptstyle\surd$               &             & $\scriptstyle\surd$           & 75.33\%                         & \textbf{72.02\%}             & 72.99\%                           \\
                                      & $\scriptstyle\surd$               & $\scriptstyle\surd$           &             & 79.10\%                         & 70.50\%                      & 75.80\%                           \\
                                      & $\scriptstyle\surd$               & $\scriptstyle\surd$           & $\scriptstyle\surd$           & \textbf{80.70\%}                & 64.93\%                      & \textbf{77.91\%}                  \\ \hline
\multirow{4}{*}{Southwest France}     &                 & $\scriptstyle\surd$           & $\scriptstyle\surd$           & 84.68\%                         & 75.77\%                      & 79.37\%                           \\
                                      & $\scriptstyle\surd$               &             & $\scriptstyle\surd$           & 85.73\%                         & 81.48\%                      & 79.89\%                           \\
                                      & $\scriptstyle\surd$               & $\scriptstyle\surd$           &             & 77.28\%                         & \textbf{90.66\%}             & 73.49\%                           \\
                                      & $\scriptstyle\surd$               & $\scriptstyle\surd$           & $\scriptstyle\surd$           & \textbf{87.56\%}                & 73.99\%                      & \textbf{80.50\%}                  \\ \hline
\multirow{4}{*}{Kansas, USA}    &                 & $\scriptstyle\surd$           & $\scriptstyle\surd$           & 91.49\%                         & \textbf{82.30\%}             & 88.13\%                           \\
                                      & $\scriptstyle\surd$               &             & $\scriptstyle\surd$           & 86.87\%                         & 81.12\%                      & 85.48\%                           \\
                                      & $\scriptstyle\surd$               & $\scriptstyle\surd$           &             & 90.40\%                         & 82.26\%                      & 88.21\%                           \\
                                      & $\scriptstyle\surd$               & $\scriptstyle\surd$           & $\scriptstyle\surd$           & \textbf{91.65\%}                & 72.85\%                      & \textbf{88.36\%}                  \\ \hline
\end{tabular*}
}
\end{table}

In the proposed framework, the use of different GLC products as inputs results in diverse high-quality labels, which directly affect the supervised part of the learning signal. To analyze the impacts of these inputs, we used different combinations of ESA, Esri, and Dyworld as inputs, and calculated the number and accuracy of the obtained high-quality labels. The results in Table 5 demonstrate that the combination of all three products yields the best performance across all the study areas, with the label accuracy showing the closest correlation with the final prediction outcome. In the Hunan study areas, although the combination of the Dyworld and Esri gets the highest label ratio of 72.02\%, its prediction F1-score is the lowest at 72.99\%, influenced by its lowest label accuracy of 75.33\%. In the study areas of Southwest France, the combination of Dyworld and ESA demonstrates the highest label ratio of 90.66\%, but with the lowest label accuracy of 77.28\%, which leads to the lowest prediction F1-score of 73.49\%. In the Kansas study areas, the combination of all three GLC products has a label ratio of 72.85\%, but it has higher prediction accuracy than the rest combinations, because of its label accuracy of 91.65\%. The reason is that our framework uses unsupervised learning signals to incorporate the samples without high-quality labels into the model learning process, reducing the model’s high dependence on the label number.

\begin{figure}[ht]
	\centering
		\includegraphics[width=0.95\linewidth]{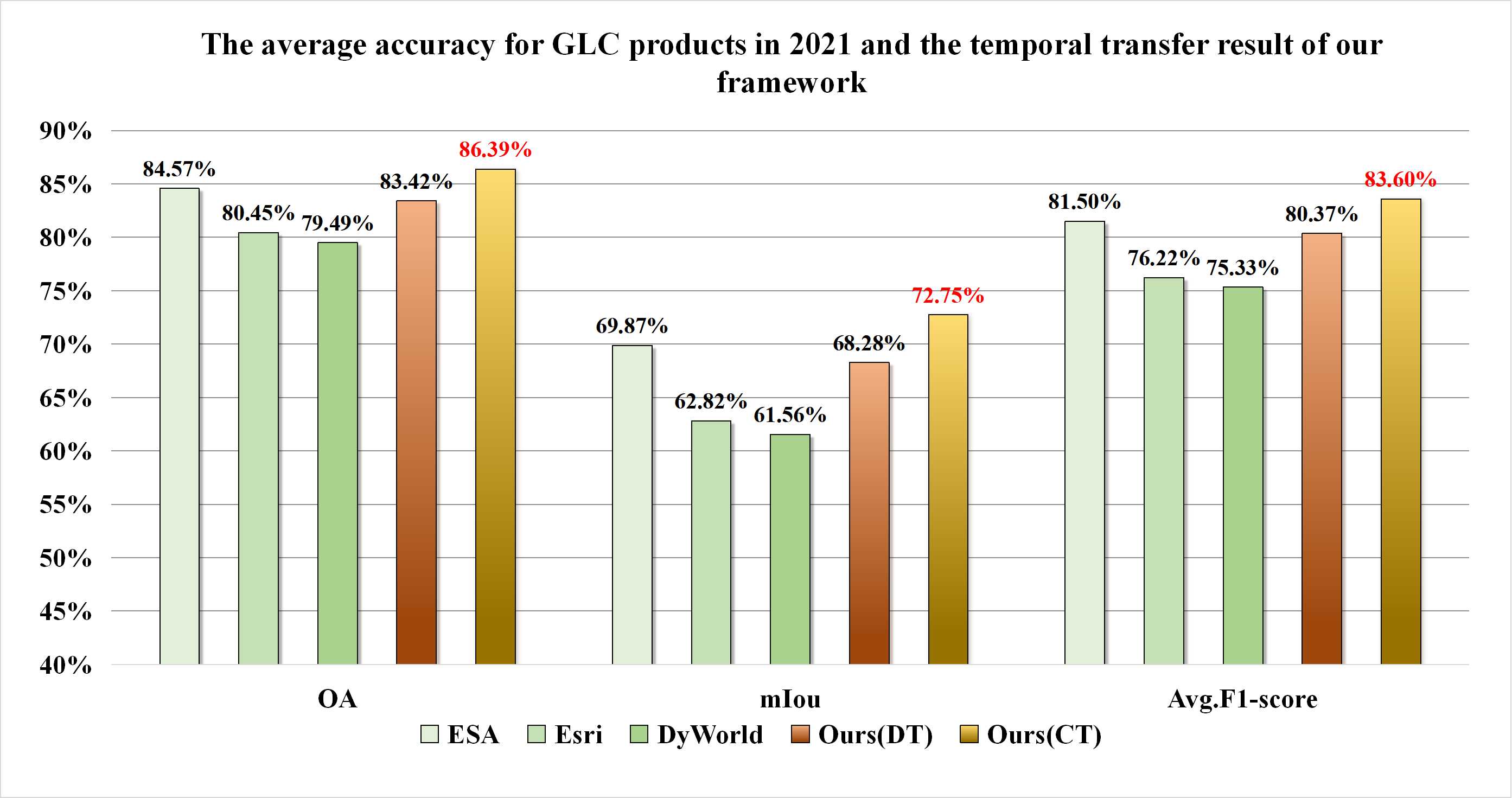}
	  \caption{The average accuracy of our framework following Direct Transfer (DT), Continue Training (CT), and the accuracy of other GLC products in 2021 across three study areas.}\label{fig12}
\end{figure}

\begin{figure}[ht]
	\centering
		\includegraphics[width=0.9\linewidth]{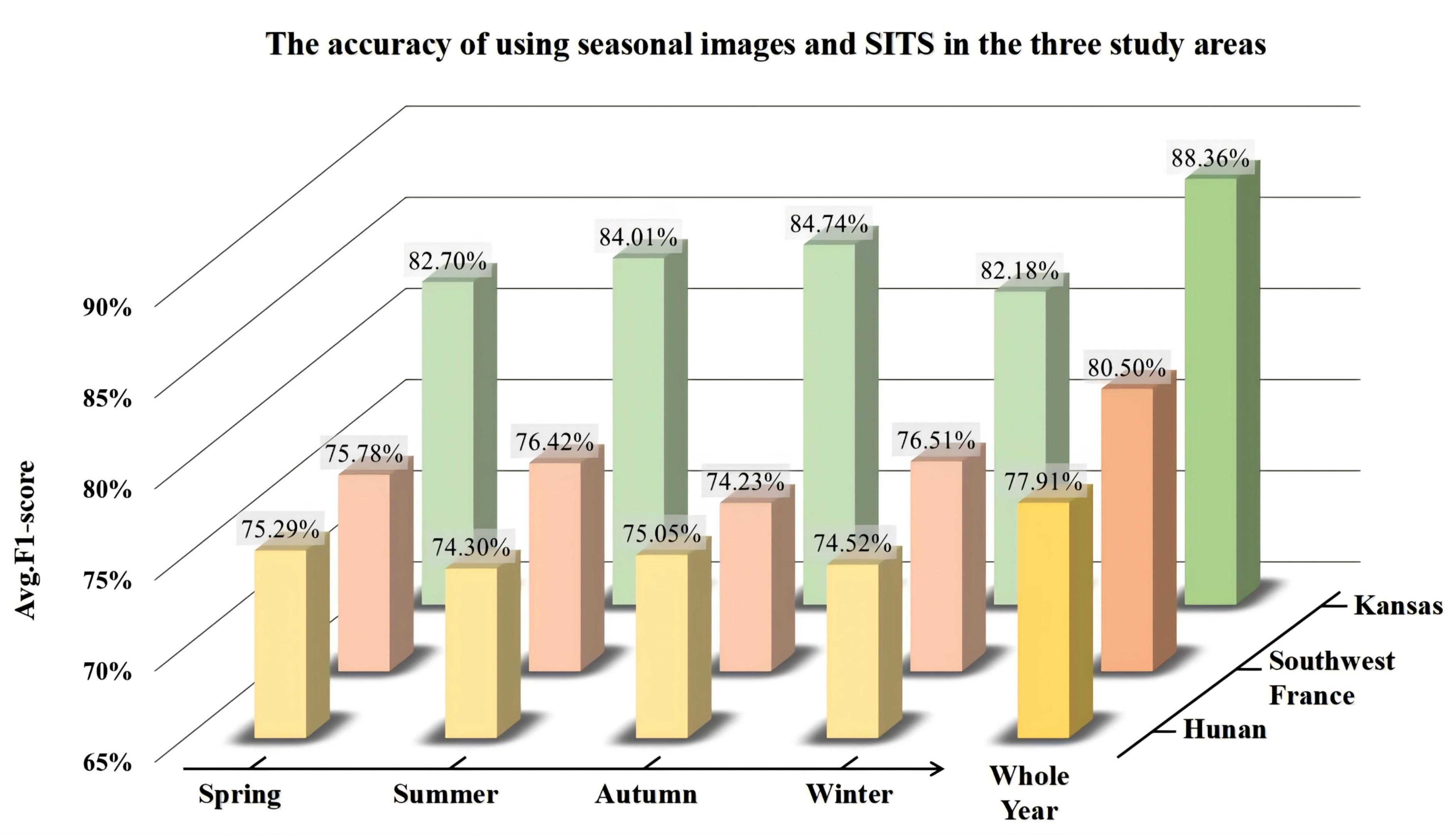}
	  \caption{The classification accuracy of using seasonal composite images and whole year SITS in the three study areas.}\label{fig13}
\end{figure}

\subsection{Assessment of temporal generalizability}\label{Section 5.2}

\medskip
A key challenge in large-scale cropland mapping is the generalization capacity of the framework. Influenced by different agrosystems and climates, the cropland exhibits various phenological attributes on SITS for different spatial and temporal coverage. Section 4.3 provides a comprehensive analysis of the model’s spatial generalization capability in the three study areas, but the model’s temporal generalization ability remains unclear. To address this, we randomly selected 1,000 samples of changed cropland from each of the three study areas, each sample with a size of 256*256 pixels. For those samples, we collected the corresponding SITS and GLC products for the year 2021, while manually labeling them with the assistance of Google Earth images. Furthermore, we designed two sets of experiments to demonstrate the temporal generalizability of our method: (1) Direct Transfer (DT): The model trained on the 2020 data was directly employed on the 2021 data without any modification; and (2) Continue Training (CT): The model trained on the 2020 data was further trained using 2021 data and then applied to the 2021 data.

\par
As shown in \cref{fig12}, the average accuracy of our framework after the DT operation does not exceed that of the best GLC products. This is due to the changing phenological feature of cropland influenced by varying climatic conditions and planting patterns between years, which results in the models trained on the 2020 data being unsuitable for the 2021 data. However, our method does not require any manual labeling cost, which facilitates the incorporation of new data for continued model training. This allows the model to progressively enhance the ability to recognize the croplands with different phenological features. Therefore, our framework after the CT operation achieved superior and more stable performance compared to other products. It outperformed the best-performing GLC product by 1.82\%, 2.87\%, and 2.10\% in OA, mIoU, and F1- scores, respectively across all study areas.
\par
This result illustrates the limitations of our method regarding temporal generalizability when directly transferred. However, these limitations can be addressed by further training the model using available data without labeling costs.

\begin{figure}[ht]
	\centering
		\includegraphics[width=0.85\linewidth]{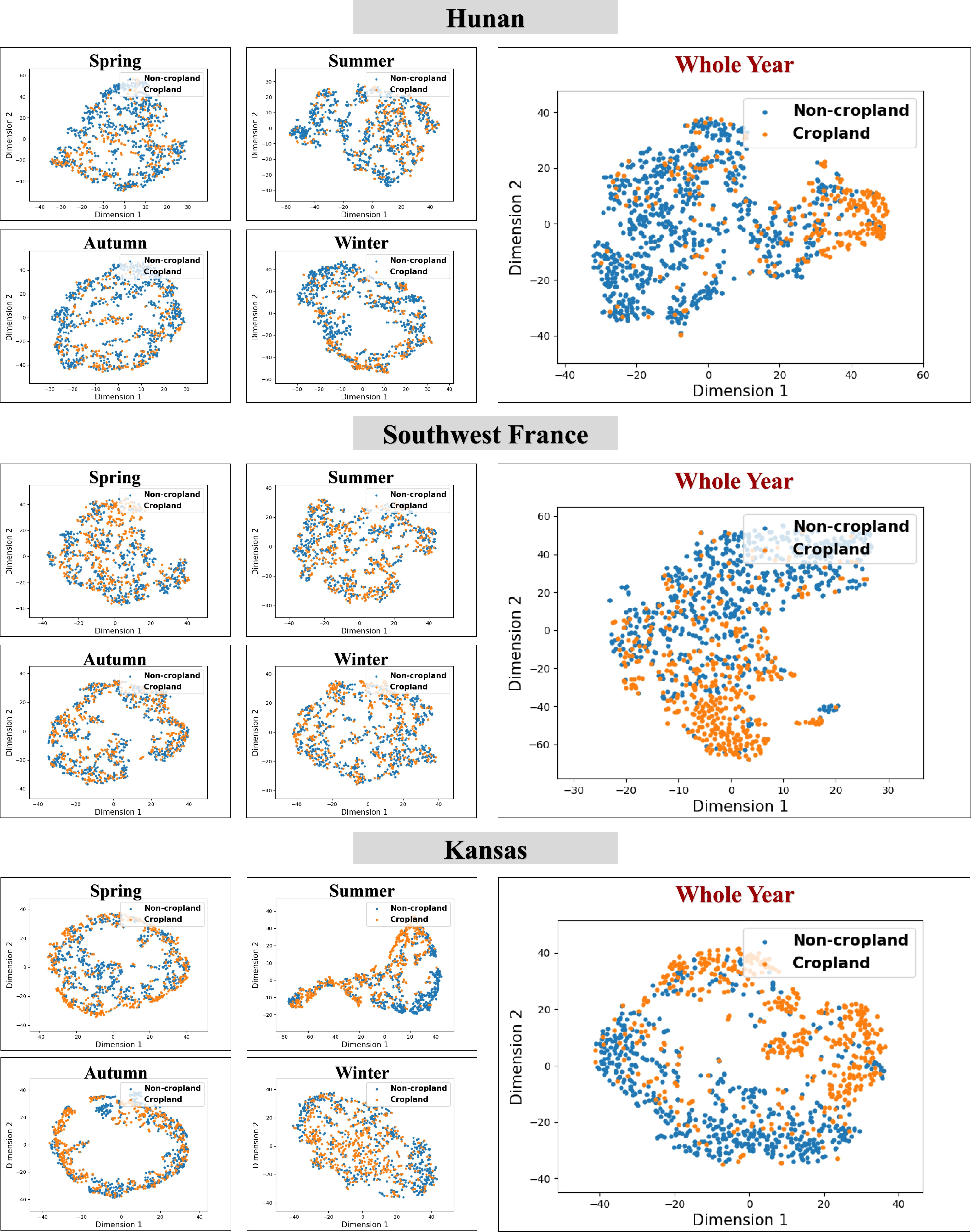}
	  \caption{The t-SNE visualization results of seasonal composite images and whole year SITS in the three study areas.}\label{fig14}
\end{figure}

\subsection{Exploring the Imperative Role of Time-series Information}\label{Section 5.3}

\medskip
To analyze the necessity of incorporating temporal information, we used temporal data from each season to generate the seasonal composite images. These images were used as input to train the model, following the proposed framework but excluding the temporal encoding part. As shown in \cref{fig13}, when compared with the highest accuracy obtained by using only the seasonal composite images, the extraction results from the entire year’s SITS exhibit improvement of 3.47\%, 5.22\%, and 4.26\% in Hunan, Southwest France, and Kansas, respectively. The main reasons for this phenomenon are as follows:
\par
Considering cropland as a whole, it displays significant visual variations across different periods due to its planting status. This poses challenges for accurately delineating the extent of cropland, as it can resemble other land covers during some periods. For instance, in the Hunan area, where rice cultivation is prevalent, the planting season falls in early summer and harvesting occurs in autumn. Thus, the cropland exhibits similar visual features to bare land in spring and winter, leading to relatively low extraction accuracy during these two seasons:
\par
Inside one cropland, there may be multiple crop types with different phenological patterns, which leads to large intra-class diversity in a specific period. In this case, the models only relying on textural, spectral, and spatial features in a single time phase can hardly recognize every part of the cropland. For instance, in Hunan, oilseed rape is often planted in the autumn, coinciding with the harvest of rice. Although the harvested rice field can be easily distinguished by its unique textural feature, oilseed rape may visually resemble other vegetation covers like shrubs or lawns during that period. This similarity can lead to misidentification of oilseed rape as other vegetation types. As illustrated in \cref{fig13}, the extraction result in Hunan areas remains low during the autumn seasons.
\par
Therefore, we extracted phenological features by integrating dense SITS and employed them to enhance the separability of cropland within the model’s feature space, thereby augmenting the completeness of cropland extraction. To further explore the benefit of the incorporation of multi-temporal information, we employed the t-distributed stochastic neighbor embedding (t-SNE) method to visualize intermediate feature maps of the models trained with seasonal composite images and whole-year SITS. As shown in \cref{fig14}, the intermediate feature maps derived from models using whole-year SITS demonstrated superior separability across all three study areas compared to those using seasonal composite images. Specifically, regarding the entirety of cropland, the inclusion of multi-temporal information can help the model to better distinguish between cropland and other land covers, thereby reducing feature confusion related to planting status during specific periods. In terms of intra-cropland, multi-temporal information enables the model to reduce intra-class feature dissimilarity within the croplands, facilitating the recognition of diverse croplands encompassing various crop types simultaneously.

\subsection{Robustness Analysis of Cloud Cover Scenarios}\label{Section 5.4}

\medskip

In practical application scenarios, SITS often suffers from information loss caused by cloud cover, which limits the model’s ability to extract complete spatial and temporal features. To assess the robustness of the proposed framework in cloud cover scenarios, we conducted experiments simulating cloud cover effects in both temporal and spatial dimensions. Considering real-world scenarios \citep{coluzziFirstAssessmentSentinel22018}, we designed the experiment as follows: in the spatial dimension, we added cloud masks of various sizes to the images, ranging from 0.00\% to 40.00\% in increments of 10.00\%. In the temporal dimension, we simulated data-missing situations by randomly dropping images. The dropping rate was set from 0\% to 83.33\% at intervals of 8.33\%, which simulates the situation where images from one to ten months were missing. Given that data loss in both spatial and temporal dimensions often occurs at the same time, we simultaneously imposed the masking and dropping operation in the simulation experiment. 

\begin{figure}[ht]
	\centering
		\includegraphics[width=1\linewidth]{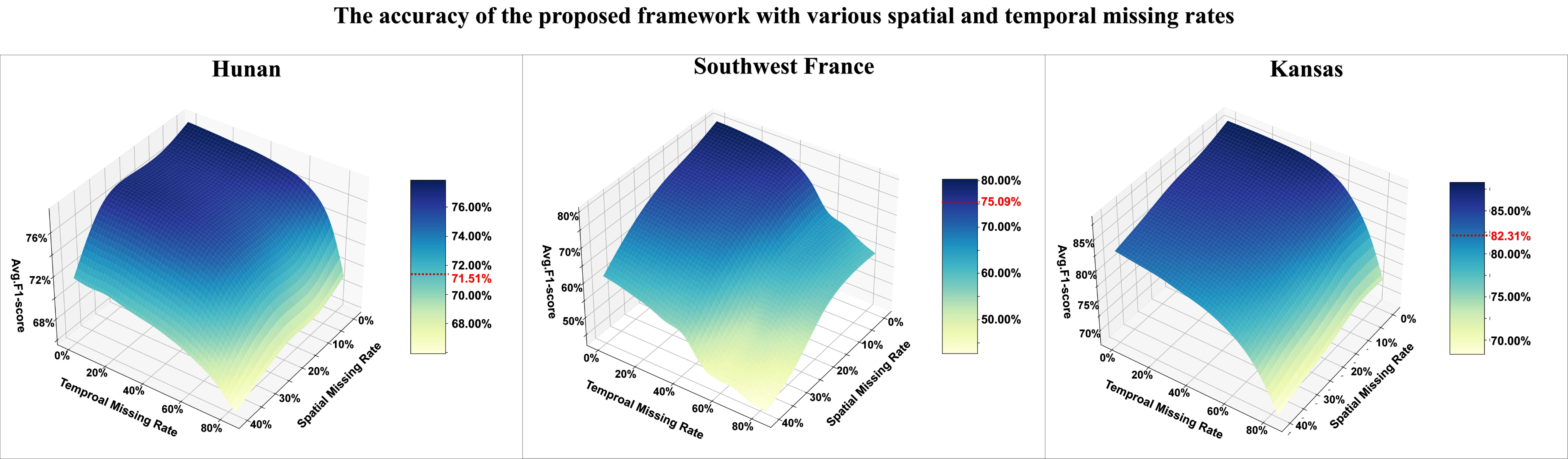}
	  \caption{The accuracy of the proposed framework in situations of various spatial and temporal missing rates. The red line in the color bar indicates the average accuracy of three GLC products in the corresponding study area.}\label{fig15}
\end{figure}

\par
The results in \cref{fig15} show that our framework exhibits considerable robustness to the data loss caused by cloud cover in all three study areas. In the Hunan study area, compared to the Avg.F1-score of multiple GLC products at 71.51\%, our framework still exhibits better performance even with a spatial missing rate of 30\% and a temporal missing rate of 66.67\%. In the Southwest France study area, compared with the Avg.F1-score of 75.09\% for multiple GLC products, our framework shows better performance under the situation of 10.00\% spatial missing rate and 33.33\% temporal missing rate. In the Kansas study area, compared with the Avg.F1-score of 82.31\% for multiple GLC products, our framework has better performance under the situation of 20.00\% spatial missing rate and 50.00\% temporal missing rate. These results demonstrate the strong feasibility of our framework in real-world data-missing scenarios. The exceptional performance in the Hunan study area may be attributed to the training data affected by distinctive climatic conditions. The local model has been already exposed to the data-missing situation in the training process, allowing the model to gain better adaptability in this case.

\section{Conclusion}\label{Section 6}
In this study, we proposed a weakly supervised framework for large-scale cropland mapping using multi-temporal information. The framework uses the labels from existing GLC products and dense SITS to capture the diverse temporal features of cropland influenced by crop phenology and human agricultural activities, without the need for manual labeling. The approach enables the model to effectively utilize the information in low-quality labeled samples, while avoiding over-fitting the residual errors in high-quality labeled samples. In the experiments across three study areas, the proposed framework demonstrated superiority over GLC products and outperformed the traditional methods that rely on discard and re-correct methods. Furthermore, we investigated the effects of input setting and the temporal generalizability of the proposed framework, while exploring the necessity of using multi-temporal information and the robustness of our framework in a cloud cover scenario. Further efforts can be made to enhance the efficiency of the proposed framework and reinforce the robustness of the model in real-world scenarios with missing information.

\section*{Acknowledgements}
The work presented in this paper was supported by the National Natural Science Foundation of China (No. 42171376); The Distinguished Young Scholars under Grant 2022JJ10072; and the Open Fund of Xiangjiang Laboratory under Grant 22XJ03007.




\clearpage 







\printcredits

\bibliographystyle{elsarticle-harv}

\bibliography{ECA_10_20}

\begin{thebibliography}{66}
\expandafter\ifx\csname natexlab\endcsname\relax\def\natexlab#1{#1}\fi
\providecommand{\url}[1]{\texttt{#1}}
\providecommand{\href}[2]{#2}
\providecommand{\path}[1]{#1}
\providecommand{\DOIprefix}{doi:}
\providecommand{\ArXivprefix}{arXiv:}
\providecommand{\URLprefix}{URL: }
\providecommand{\Pubmedprefix}{pmid:}
\providecommand{\doi}[1]{\href{http://dx.doi.org/#1}{\path{#1}}}
\providecommand{\Pubmed}[1]{\href{pmid:#1}{\path{#1}}}
\providecommand{\bibinfo}[2]{#2}
\ifx\xfnm\relax \def\xfnm[#1]{\unskip,\space#1}\fi
\bibitem[{Amani et~al.(2020)Amani, Ghorbanian, Ahmadi, Kakooei, Moghimi, Mirmazloumi, Moghaddam, Mahdavi, Ghahremanloo, Parsian, Wu and Brisco}]{amaniGoogleEarthEngine2020a}
\bibinfo{author}{Amani, M.}, \bibinfo{author}{Ghorbanian, A.}, \bibinfo{author}{Ahmadi, S.A.}, \bibinfo{author}{Kakooei, M.}, \bibinfo{author}{Moghimi, A.}, \bibinfo{author}{Mirmazloumi, S.M.}, \bibinfo{author}{Moghaddam, S.H.A.}, \bibinfo{author}{Mahdavi, S.}, \bibinfo{author}{Ghahremanloo, M.}, \bibinfo{author}{Parsian, S.}, \bibinfo{author}{Wu, Q.}, \bibinfo{author}{Brisco, B.}, \bibinfo{year}{2020}.
\newblock \bibinfo{title}{Google {{Earth Engine Cloud Computing Platform}} for {{Remote Sensing Big Data Applications}}: {{A Comprehensive Review}}}.
\newblock \bibinfo{journal}{IEEE J. Sel. Top. Appl. Earth Observations Remote Sensing} \bibinfo{volume}{13}, \bibinfo{pages}{5326--5350}.
\newblock \bibinfo{note}{\url{https://ieeexplore.ieee.org/document/9184118/}}.
\bibitem[{Belgiu and Csillik(2018)}]{belgiuSentinel2CroplandMapping2018d}
\bibinfo{author}{Belgiu, M.}, \bibinfo{author}{Csillik, O.}, \bibinfo{year}{2018}.
\newblock \bibinfo{title}{Sentinel-2 cropland mapping using pixel-based and object-based time-weighted dynamic time warping analysis}.
\newblock \bibinfo{journal}{Remote Sensing of Environment} \bibinfo{volume}{204}, \bibinfo{pages}{509--523}.
\newblock \bibinfo{note}{\url{https://linkinghub.elsevier.com/retrieve/pii/S0034425717304686}}.
\bibitem[{Brown et~al.(2022a)Brown, Brumby, {Guzder-Williams}, Birch, Hyde, Mazzariello, Czerwinski, Pasquarella, Haertel, Ilyushchenko, Schwehr, Weisse, Stolle, Hanson, Guinan, Moore and Tait}]{brownDynamicWorldRealtime2022b}
\bibinfo{author}{Brown, C.F.}, \bibinfo{author}{Brumby, S.P.}, \bibinfo{author}{{Guzder-Williams}, B.}, \bibinfo{author}{Birch, T.}, \bibinfo{author}{Hyde, S.B.}, \bibinfo{author}{Mazzariello, J.}, \bibinfo{author}{Czerwinski, W.}, \bibinfo{author}{Pasquarella, V.J.}, \bibinfo{author}{Haertel, R.}, \bibinfo{author}{Ilyushchenko, S.}, \bibinfo{author}{Schwehr, K.}, \bibinfo{author}{Weisse, M.}, \bibinfo{author}{Stolle, F.}, \bibinfo{author}{Hanson, C.}, \bibinfo{author}{Guinan, O.}, \bibinfo{author}{Moore, R.}, \bibinfo{author}{Tait, A.M.}, \bibinfo{year}{2022}a.
\newblock \bibinfo{title}{Dynamic {{World}}, {{Near}} real-time global 10 m land use land cover mapping}.
\newblock \bibinfo{journal}{Sci Data} \bibinfo{volume}{9}, \bibinfo{pages}{251}.
\newblock \bibinfo{note}{\url{https://www.nature.com/articles/s41597-022-01307-4}}.
\bibitem[{Brown et~al.(2022b)Brown, Brumby, Guzder-Williams, Birch, Hyde, Mazzariello, Czerwinski, Pasquarella, Haertel, Ilyushchenko et~al.}]{brown2022dynamic}
\bibinfo{author}{Brown, C.F.}, \bibinfo{author}{Brumby, S.P.}, \bibinfo{author}{Guzder-Williams, B.}, \bibinfo{author}{Birch, T.}, \bibinfo{author}{Hyde, S.B.}, \bibinfo{author}{Mazzariello, J.}, \bibinfo{author}{Czerwinski, W.}, \bibinfo{author}{Pasquarella, V.J.}, \bibinfo{author}{Haertel, R.}, \bibinfo{author}{Ilyushchenko, S.}, et~al., \bibinfo{year}{2022}b.
\newblock \bibinfo{title}{Dynamic world, near real-time global 10 m land use land cover mapping}.
\newblock \bibinfo{journal}{Scientific Data} \bibinfo{volume}{9}, \bibinfo{pages}{251}.
\bibitem[{Calvao and Pessoa(2015)}]{calvaoRemoteSensingFood2015a}
\bibinfo{author}{Calvao, T.}, \bibinfo{author}{Pessoa, M.}, \bibinfo{year}{2015}.
\newblock \bibinfo{title}{Remote sensing in food production-a review}.
\newblock \bibinfo{journal}{Emir. J. Food Agric} \bibinfo{volume}{27}, \bibinfo{pages}{138}.
\newblock \bibinfo{note}{\url{http://www.ejfa.me/index.php/journal/article/view/652}}.
\bibitem[{Chen et~al.(2018)Chen, Zhu, Papandreou, Schroff and Adam}]{chenEncoderDecoderAtrousSeparable2018a}
\bibinfo{author}{Chen, L.C.}, \bibinfo{author}{Zhu, Y.}, \bibinfo{author}{Papandreou, G.}, \bibinfo{author}{Schroff, F.}, \bibinfo{author}{Adam, H.}, \bibinfo{year}{2018}.
\newblock \bibinfo{title}{Encoder-{{Decoder}} with {{Atrous Separable Convolution}} for {{Semantic Image Segmentation}}}, in: \bibinfo{editor}{Ferrari, V.}, \bibinfo{editor}{Hebert, M.}, \bibinfo{editor}{Sminchisescu, C.}, \bibinfo{editor}{Weiss, Y.} (Eds.), \bibinfo{booktitle}{Computer {{Vision}} {\textendash} {{ECCV}} 2018}. \bibinfo{publisher}{{Springer International Publishing}}, \bibinfo{address}{{Cham}}. volume \bibinfo{volume}{11211}, pp. \bibinfo{pages}{833--851}.
\newblock \bibinfo{note}{\url{https://link.springer.com/10.1007/978-3-030-01234-2_49}}.
\bibitem[{Chen et~al.(2023)Chen, Zhang, Cui, Li, Hou, Ma, Li, Li and Wang}]{chenNovelWeaklySupervised2023b}
\bibinfo{author}{Chen, Y.}, \bibinfo{author}{Zhang, G.}, \bibinfo{author}{Cui, H.}, \bibinfo{author}{Li, X.}, \bibinfo{author}{Hou, S.}, \bibinfo{author}{Ma, J.}, \bibinfo{author}{Li, Z.}, \bibinfo{author}{Li, H.}, \bibinfo{author}{Wang, H.}, \bibinfo{year}{2023}.
\newblock \bibinfo{title}{A novel weakly supervised semantic segmentation framework to improve the resolution of land cover product}.
\newblock \bibinfo{journal}{ISPRS Journal of Photogrammetry and Remote Sensing} \bibinfo{volume}{196}, \bibinfo{pages}{73--92}.
\newblock \bibinfo{note}{\url{https://linkinghub.elsevier.com/retrieve/pii/S0924271622003422}}.
\bibitem[{Chi et~al.(2016)Chi, Plaza, Benediktsson, Sun, Shen and Zhu}]{chiBigDataRemote2016a}
\bibinfo{author}{Chi, M.}, \bibinfo{author}{Plaza, A.}, \bibinfo{author}{Benediktsson, J.A.}, \bibinfo{author}{Sun, Z.}, \bibinfo{author}{Shen, J.}, \bibinfo{author}{Zhu, Y.}, \bibinfo{year}{2016}.
\newblock \bibinfo{title}{Big {{Data}} for {{Remote Sensing}}: {{Challenges}} and {{Opportunities}}}.
\newblock \bibinfo{journal}{Proc. IEEE} \bibinfo{volume}{104}, \bibinfo{pages}{2207--2219}.
\newblock \bibinfo{note}{\url{https://ieeexplore.ieee.org/document/7565634/}}.
\bibitem[{Coluzzi et~al.(2018)Coluzzi, Imbrenda, Lanfredi and Simoniello}]{coluzziFirstAssessmentSentinel22018}
\bibinfo{author}{Coluzzi, R.}, \bibinfo{author}{Imbrenda, V.}, \bibinfo{author}{Lanfredi, M.}, \bibinfo{author}{Simoniello, T.}, \bibinfo{year}{2018}.
\newblock \bibinfo{title}{A first assessment of the {{Sentinel-2 Level}} 1-{{C}} cloud mask product to support informed surface analyses}.
\newblock \bibinfo{journal}{Remote Sensing of Environment} \bibinfo{volume}{217}, \bibinfo{pages}{426--443}.
\newblock \bibinfo{note}{\url{https://linkinghub.elsevier.com/retrieve/pii/S0034425718303742}}.
\bibitem[{{Copernicus Climate Change Service}(2019)}]{copernicusclimatechangeserviceLandCoverClassification2019a}
\bibinfo{author}{{Copernicus Climate Change Service}}, \bibinfo{year}{2019}.
\newblock \bibinfo{title}{Land cover classification gridded maps from 1992 to present derived from satellite observations}.
\newblock \bibinfo{howpublished}{\url{https://cds.climate.copernicus.eu/doi/10.24381/cds.006f2c9a}}.
\bibitem[{Defourny et~al.(2019)Defourny, Bontemps, Bellemans, Cara, Dedieu, Guzzonato, Hagolle, Inglada, Nicola, Rabaute, Savinaud, Udroiu, Valero, B{\'e}gu{\'e}, Dejoux, El~Harti, Ezzahar, Kussul, Labbassi, Lebourgeois, Miao, Newby, Nyamugama, Salh, Shelestov, Simonneaux, Traore, Traore and Koetz}]{defournyRealtimeAgricultureMonitoring2019}
\bibinfo{author}{Defourny, P.}, \bibinfo{author}{Bontemps, S.}, \bibinfo{author}{Bellemans, N.}, \bibinfo{author}{Cara, C.}, \bibinfo{author}{Dedieu, G.}, \bibinfo{author}{Guzzonato, E.}, \bibinfo{author}{Hagolle, O.}, \bibinfo{author}{Inglada, J.}, \bibinfo{author}{Nicola, L.}, \bibinfo{author}{Rabaute, T.}, \bibinfo{author}{Savinaud, M.}, \bibinfo{author}{Udroiu, C.}, \bibinfo{author}{Valero, S.}, \bibinfo{author}{B{\'e}gu{\'e}, A.}, \bibinfo{author}{Dejoux, J.F.}, \bibinfo{author}{El~Harti, A.}, \bibinfo{author}{Ezzahar, J.}, \bibinfo{author}{Kussul, N.}, \bibinfo{author}{Labbassi, K.}, \bibinfo{author}{Lebourgeois, V.}, \bibinfo{author}{Miao, Z.}, \bibinfo{author}{Newby, T.}, \bibinfo{author}{Nyamugama, A.}, \bibinfo{author}{Salh, N.}, \bibinfo{author}{Shelestov, A.}, \bibinfo{author}{Simonneaux, V.}, \bibinfo{author}{Traore, P.S.}, \bibinfo{author}{Traore, S.S.}, \bibinfo{author}{Koetz, B.}, \bibinfo{year}{2019}.
\newblock \bibinfo{title}{Near real-time agriculture monitoring at national scale at parcel resolution: {{Performance}} assessment of the {{Sen2-Agri}} automated system in various cropping systems around the world}.
\newblock \bibinfo{journal}{Remote Sensing of Environment} \bibinfo{volume}{221}, \bibinfo{pages}{551--568}.
\newblock \bibinfo{note}{\url{https://linkinghub.elsevier.com/retrieve/pii/S0034425718305145}}.
\bibitem[{Deng et~al.(2021)Deng, Xu and Huang}]{deng2021cnns}
\bibinfo{author}{Deng, P.}, \bibinfo{author}{Xu, K.}, \bibinfo{author}{Huang, H.}, \bibinfo{year}{2021}.
\newblock \bibinfo{title}{When cnns meet vision transformer: A joint framework for remote sensing scene classification}.
\newblock \bibinfo{journal}{IEEE Geoscience and Remote Sensing Letters} \bibinfo{volume}{19}, \bibinfo{pages}{1--5}.
\bibitem[{Do~Nascimento~Bendini et~al.(2019)Do~Nascimento~Bendini, Garcia~Fonseca, Schwieder, Sehn~K{\"o}rting, Rufin, Del Arco~Sanches, Leit{\~a}o and Hostert}]{donascimentobendiniDetailedAgriculturalLand2019a}
\bibinfo{author}{Do~Nascimento~Bendini, H.}, \bibinfo{author}{Garcia~Fonseca, L.M.}, \bibinfo{author}{Schwieder, M.}, \bibinfo{author}{Sehn~K{\"o}rting, T.}, \bibinfo{author}{Rufin, P.}, \bibinfo{author}{Del Arco~Sanches, I.}, \bibinfo{author}{Leit{\~a}o, P.J.}, \bibinfo{author}{Hostert, P.}, \bibinfo{year}{2019}.
\newblock \bibinfo{title}{Detailed agricultural land classification in the {{Brazilian}} cerrado based on phenological information from dense satellite image time series}.
\newblock \bibinfo{journal}{International Journal of Applied Earth Observation and Geoinformation} \bibinfo{volume}{82}, \bibinfo{pages}{101872}.
\newblock \bibinfo{note}{\url{https://linkinghub.elsevier.com/retrieve/pii/S0303243418308961}}.
\bibitem[{FAO.(2010)}]{fao2010world}
\bibinfo{author}{FAO.}, \bibinfo{year}{2010}.
\newblock \bibinfo{title}{World programme for the census of agriculture}.
\bibitem[{Fare~Garnot and Landrieu(2021)}]{faregarnotPanopticSegmentationSatellite2021b}
\bibinfo{author}{Fare~Garnot, V.S.}, \bibinfo{author}{Landrieu, L.}, \bibinfo{year}{2021}.
\newblock \bibinfo{title}{Panoptic {{Segmentation}} of {{Satellite Image Time Series}} with {{Convolutional Temporal Attention Networks}}}, in: \bibinfo{booktitle}{2021 {{IEEECVF Int}}. {{Conf}}. {{Comput}}. {{Vis}}. {{ICCV}}}, \bibinfo{publisher}{{IEEE}}, \bibinfo{address}{{Montreal, QC, Canada}}. pp. \bibinfo{pages}{4852--4861}.
\newblock \bibinfo{note}{\url{https://ieeexplore.ieee.org/document/9711189/}}.
\bibitem[{{Food and Agriculture Organization of the United Nations}(2005)}]{FoodandAgricultureOrganizationoftheUnitedNations_SystemIntegratedAgricultural_2005}
\bibinfo{editor}{{Food and Agriculture Organization of the United Nations}} (Ed.), \bibinfo{year}{2005}.
\newblock \bibinfo{title}{A System of Integrated Agricultural Censuses and Surveys}.
\newblock Number~\bibinfo{number}{11} in \bibinfo{series}{{{FAO}} Statistical Development Series}, \bibinfo{publisher}{{Food and Agriculture Organization of the United Nations}}, \bibinfo{address}{Rome}.
\bibitem[{Friedl and {Sulla-Menashe}(2019)}]{friedlMCD12Q1MODISTerra2019}
\bibinfo{author}{Friedl, M.}, \bibinfo{author}{{Sulla-Menashe}, D.}, \bibinfo{year}{2019}.
\newblock \bibinfo{title}{{{MCD12Q1 MODIS}}/{{Terra}}+{{Aqua Land Cover Type Yearly L3 Global}} 500m {{SIN Grid V006}}}.
\newblock \bibinfo{howpublished}{\url{https://lpdaac.usgs.gov/products/mcd12q1v006/}}.
\bibitem[{Garnot and Landrieu(2020)}]{garnotLightweightTemporalSelfattention2020b}
\bibinfo{author}{Garnot, V.S.F.}, \bibinfo{author}{Landrieu, L.}, \bibinfo{year}{2020}.
\newblock \bibinfo{title}{Lightweight {{Temporal Self-attention}} for {{Classifying Satellite Images Time Series}}}, in: \bibinfo{editor}{Lemaire, V.}, \bibinfo{editor}{Malinowski, S.}, \bibinfo{editor}{Bagnall, A.}, \bibinfo{editor}{Guyet, T.}, \bibinfo{editor}{Tavenard, R.}, \bibinfo{editor}{Ifrim, G.} (Eds.), \bibinfo{booktitle}{Advanced {{Analytics}} and {{Learning}} on {{Temporal Data}}}. \bibinfo{publisher}{{Springer International Publishing}}, \bibinfo{address}{{Cham}}. volume \bibinfo{volume}{12588}, pp. \bibinfo{pages}{171--181}.
\newblock \bibinfo{note}{\url{http://link.springer.com/10.1007/978-3-030-65742-0_12}}.
\bibitem[{Gong et~al.(2013)Gong, Wang, Yu, Zhao, Zhao, Liang, Niu, Huang, Fu, Liu, Li, Li, Fu, Liu, Xu, Wang, Cheng, Hu, Yao, Zhang, Zhu, Zhao, Zhang, Zheng, Ji, Zhang, Chen, Yan, Guo, Yu, Wang, Liu, Shi, Zhu, Chen, Yang, Tang, Xu, Giri, Clinton, Zhu, Chen and Chen}]{gongFinerResolutionObservation2013a}
\bibinfo{author}{Gong, P.}, \bibinfo{author}{Wang, J.}, \bibinfo{author}{Yu, L.}, \bibinfo{author}{Zhao, Y.}, \bibinfo{author}{Zhao, Y.}, \bibinfo{author}{Liang, L.}, \bibinfo{author}{Niu, Z.}, \bibinfo{author}{Huang, X.}, \bibinfo{author}{Fu, H.}, \bibinfo{author}{Liu, S.}, \bibinfo{author}{Li, C.}, \bibinfo{author}{Li, X.}, \bibinfo{author}{Fu, W.}, \bibinfo{author}{Liu, C.}, \bibinfo{author}{Xu, Y.}, \bibinfo{author}{Wang, X.}, \bibinfo{author}{Cheng, Q.}, \bibinfo{author}{Hu, L.}, \bibinfo{author}{Yao, W.}, \bibinfo{author}{Zhang, H.}, \bibinfo{author}{Zhu, P.}, \bibinfo{author}{Zhao, Z.}, \bibinfo{author}{Zhang, H.}, \bibinfo{author}{Zheng, Y.}, \bibinfo{author}{Ji, L.}, \bibinfo{author}{Zhang, Y.}, \bibinfo{author}{Chen, H.}, \bibinfo{author}{Yan, A.}, \bibinfo{author}{Guo, J.}, \bibinfo{author}{Yu, L.}, \bibinfo{author}{Wang, L.}, \bibinfo{author}{Liu, X.}, \bibinfo{author}{Shi, T.}, \bibinfo{author}{Zhu, M.}, \bibinfo{author}{Chen, Y.}, \bibinfo{author}{Yang, G.}, \bibinfo{author}{Tang, P.},
  \bibinfo{author}{Xu, B.}, \bibinfo{author}{Giri, C.}, \bibinfo{author}{Clinton, N.}, \bibinfo{author}{Zhu, Z.}, \bibinfo{author}{Chen, J.}, \bibinfo{author}{Chen, J.}, \bibinfo{year}{2013}.
\newblock \bibinfo{title}{Finer resolution observation and monitoring of global land cover: First mapping results with {{Landsat TM}} and {{ETM}}+ data}.
\newblock \bibinfo{journal}{International Journal of Remote Sensing} \bibinfo{volume}{34}, \bibinfo{pages}{2607--2654}.
\newblock \bibinfo{note}{\url{https://www.tandfonline.com/doi/full/10.1080/01431161.2012.748992}}.
\bibitem[{Hermosilla et~al.(2022)Hermosilla, Wulder, White and Coops}]{hermosillaLandCoverClassification2022a}
\bibinfo{author}{Hermosilla, T.}, \bibinfo{author}{Wulder, M.A.}, \bibinfo{author}{White, J.C.}, \bibinfo{author}{Coops, N.C.}, \bibinfo{year}{2022}.
\newblock \bibinfo{title}{Land cover classification in an era of big and open data: {{Optimizing}} localized implementation and training data selection to improve mapping outcomes}.
\newblock \bibinfo{journal}{Remote Sensing of Environment} \bibinfo{volume}{268}, \bibinfo{pages}{112780}.
\newblock \bibinfo{note}{\url{https://linkinghub.elsevier.com/retrieve/pii/S0034425721005009}}.
\bibitem[{Hua et~al.(2021)Hua, Marcos, Mou, Zhu and Tuia}]{hua2021semantic}
\bibinfo{author}{Hua, Y.}, \bibinfo{author}{Marcos, D.}, \bibinfo{author}{Mou, L.}, \bibinfo{author}{Zhu, X.X.}, \bibinfo{author}{Tuia, D.}, \bibinfo{year}{2021}.
\newblock \bibinfo{title}{Semantic segmentation of remote sensing images with sparse annotations}.
\newblock \bibinfo{journal}{IEEE Geoscience and Remote Sensing Letters} \bibinfo{volume}{19}, \bibinfo{pages}{1--5}.
\bibitem[{Huang et~al.(2018)Huang, Chen, Yu, Huang and Gu}]{huangAgriculturalRemoteSensing2018a}
\bibinfo{author}{Huang, Y.}, \bibinfo{author}{Chen, Z.x.}, \bibinfo{author}{Yu, T.}, \bibinfo{author}{Huang, X.z.}, \bibinfo{author}{Gu, X.f.}, \bibinfo{year}{2018}.
\newblock \bibinfo{title}{Agricultural remote sensing big data: {{Management}} and applications}.
\newblock \bibinfo{journal}{Journal of Integrative Agriculture} \bibinfo{volume}{17}, \bibinfo{pages}{1915--1931}.
\newblock \bibinfo{note}{\url{https://linkinghub.elsevier.com/retrieve/pii/S2095311917618598}}.
\bibitem[{Jiang(2015)}]{jiangGeospatialAnalysisRequires2015}
\bibinfo{author}{Jiang, B.}, \bibinfo{year}{2015}.
\newblock \bibinfo{title}{Geospatial analysis requires a different way of thinking: The problem of spatial heterogeneity}.
\newblock \bibinfo{journal}{GeoJournal} \bibinfo{volume}{80}, \bibinfo{pages}{1--13}.
\newblock \bibinfo{note}{\url{http://link.springer.com/10.1007/s10708-014-9537-y}}.
\bibitem[{Karra et~al.(2021a)Karra, Kontgis, {Statman-Weil}, Mazzariello, Mathis and Brumby}]{karraGlobalLandUse2021a}
\bibinfo{author}{Karra, K.}, \bibinfo{author}{Kontgis, C.}, \bibinfo{author}{{Statman-Weil}, Z.}, \bibinfo{author}{Mazzariello, J.C.}, \bibinfo{author}{Mathis, M.}, \bibinfo{author}{Brumby, S.P.}, \bibinfo{year}{2021}a.
\newblock \bibinfo{title}{Global land use / land cover with {{Sentinel}} 2 and deep learning}, in: \bibinfo{booktitle}{2021 {{IEEE Int}}. {{Geosci}}. {{Remote Sens}}. {{Symp}}. {{IGARSS}}}, \bibinfo{publisher}{{IEEE}}, \bibinfo{address}{{Brussels, Belgium}}. pp. \bibinfo{pages}{4704--4707}.
\newblock \bibinfo{note}{\url{https://ieeexplore.ieee.org/document/9553499/}}.
\bibitem[{Karra et~al.(2021b)Karra, Kontgis, Statman-Weil, Mazzariello, Mathis and Brumby}]{karra2021global}
\bibinfo{author}{Karra, K.}, \bibinfo{author}{Kontgis, C.}, \bibinfo{author}{Statman-Weil, Z.}, \bibinfo{author}{Mazzariello, J.C.}, \bibinfo{author}{Mathis, M.}, \bibinfo{author}{Brumby, S.P.}, \bibinfo{year}{2021}b.
\newblock \bibinfo{title}{Global land use/land cover with sentinel 2 and deep learning}, in: \bibinfo{booktitle}{2021 IEEE international geoscience and remote sensing symposium IGARSS}, \bibinfo{organization}{IEEE}. pp. \bibinfo{pages}{4704--4707}.
\bibitem[{Karthikeyan et~al.(2020)Karthikeyan, Chawla and Mishra}]{karthikeyanReviewRemoteSensing2020a}
\bibinfo{author}{Karthikeyan, L.}, \bibinfo{author}{Chawla, I.}, \bibinfo{author}{Mishra, A.K.}, \bibinfo{year}{2020}.
\newblock \bibinfo{title}{A review of remote sensing applications in agriculture for food security: {{Crop}} growth and yield, irrigation, and crop losses}.
\newblock \bibinfo{journal}{Journal of Hydrology} \bibinfo{volume}{586}, \bibinfo{pages}{124905}.
\newblock \bibinfo{note}{\url{https://linkinghub.elsevier.com/retrieve/pii/S0022169420303656}}.
\bibitem[{LeCun et~al.(2015)LeCun, Bengio and Hinton}]{lecunDeepLearning2015a}
\bibinfo{author}{LeCun, Y.}, \bibinfo{author}{Bengio, Y.}, \bibinfo{author}{Hinton, G.}, \bibinfo{year}{2015}.
\newblock \bibinfo{title}{Deep learning}.
\newblock \bibinfo{journal}{Nature} \bibinfo{volume}{521}, \bibinfo{pages}{436--444}.
\newblock \bibinfo{note}{\url{https://www.nature.com/articles/nature14539}}.
\bibitem[{Lenczner et~al.(2022)Lenczner, Chan-Hon-Tong, Le~Saux, Luminari and Le~Besnerais}]{lenczner2022dial}
\bibinfo{author}{Lenczner, G.}, \bibinfo{author}{Chan-Hon-Tong, A.}, \bibinfo{author}{Le~Saux, B.}, \bibinfo{author}{Luminari, N.}, \bibinfo{author}{Le~Besnerais, G.}, \bibinfo{year}{2022}.
\newblock \bibinfo{title}{Dial: Deep interactive and active learning for semantic segmentation in remote sensing}.
\newblock \bibinfo{journal}{IEEE Journal of Selected Topics in Applied Earth Observations and Remote Sensing} \bibinfo{volume}{15}, \bibinfo{pages}{3376--3389}.
\bibitem[{Li et~al.(2021)Li, Xian, Zhou and Pengra}]{liNovelAutomaticPhenology2021b}
\bibinfo{author}{Li, C.}, \bibinfo{author}{Xian, G.}, \bibinfo{author}{Zhou, Q.}, \bibinfo{author}{Pengra, B.W.}, \bibinfo{year}{2021}.
\newblock \bibinfo{title}{A novel automatic phenology learning ({{APL}}) method of training sample selection using multiple datasets for time-series land cover mapping}.
\newblock \bibinfo{journal}{Remote Sensing of Environment} \bibinfo{volume}{266}, \bibinfo{pages}{112670}.
\newblock \bibinfo{note}{\url{https://linkinghub.elsevier.com/retrieve/pii/S0034425721003904}}.
\bibitem[{Li et~al.(2023)Li, Song, Hansen, Becker-Reshef, Adusei, Pickering, Wang, Wang, Lin, Zalles et~al.}]{li2023development}
\bibinfo{author}{Li, H.}, \bibinfo{author}{Song, X.P.}, \bibinfo{author}{Hansen, M.C.}, \bibinfo{author}{Becker-Reshef, I.}, \bibinfo{author}{Adusei, B.}, \bibinfo{author}{Pickering, J.}, \bibinfo{author}{Wang, L.}, \bibinfo{author}{Wang, L.}, \bibinfo{author}{Lin, Z.}, \bibinfo{author}{Zalles, V.}, et~al., \bibinfo{year}{2023}.
\newblock \bibinfo{title}{Development of a 10-m resolution maize and soybean map over china: Matching satellite-based crop classification with sample-based area estimation}.
\newblock \bibinfo{journal}{Remote Sensing of Environment} \bibinfo{volume}{294}, \bibinfo{pages}{113623}.
\bibitem[{Li et~al.(2019)Li, Huang and Gong}]{liDeepNeuralNetwork2019a}
\bibinfo{author}{Li, J.}, \bibinfo{author}{Huang, X.}, \bibinfo{author}{Gong, J.}, \bibinfo{year}{2019}.
\newblock \bibinfo{title}{Deep neural network for remote-sensing image interpretation: Status and perspectives}.
\newblock \bibinfo{journal}{Natl. Sci. Rev.} \bibinfo{volume}{6}, \bibinfo{pages}{1082--1086}.
\newblock \bibinfo{note}{\url{https://academic.oup.com/nsr/article/6/6/1082/5484863}}.
\bibitem[{Li and Xu(2020)}]{liCroplandDataFusion2020b}
\bibinfo{author}{Li, K.}, \bibinfo{author}{Xu, E.}, \bibinfo{year}{2020}.
\newblock \bibinfo{title}{Cropland data fusion and correction using spatial analysis techniques and the {{Google Earth Engine}}}.
\newblock \bibinfo{journal}{GIScience \& Remote Sensing} \bibinfo{volume}{57}, \bibinfo{pages}{1026--1045}.
\newblock \bibinfo{note}{\url{https://www.tandfonline.com/doi/full/10.1080/15481603.2020.1841489}}.
\bibitem[{Liu et~al.(2022)Liu, Wu, Chen, Huang, Du, Chen and Xiao}]{liuNovelImperviousSurface2022a}
\bibinfo{author}{Liu, Y.}, \bibinfo{author}{Wu, Y.}, \bibinfo{author}{Chen, Z.}, \bibinfo{author}{Huang, M.}, \bibinfo{author}{Du, W.}, \bibinfo{author}{Chen, N.}, \bibinfo{author}{Xiao, C.}, \bibinfo{year}{2022}.
\newblock \bibinfo{title}{A {{Novel Impervious Surface Extraction Method Based}} on {{Automatically Generating Training Samples From Multisource Remote Sensing Products}}: {{A Case Study}} of {{Wuhan City}}, {{China}}}.
\newblock \bibinfo{journal}{IEEE J. Sel. Top. Appl. Earth Observations Remote Sensing} \bibinfo{volume}{15}, \bibinfo{pages}{6766--6780}.
\newblock \bibinfo{note}{\url{https://ieeexplore.ieee.org/document/9854083/}}.
\bibitem[{Naboureh et~al.(2023)Naboureh, Li, Bian and Lei}]{nabourehNationalScaleLand2023a}
\bibinfo{author}{Naboureh, A.}, \bibinfo{author}{Li, A.}, \bibinfo{author}{Bian, J.}, \bibinfo{author}{Lei, G.}, \bibinfo{year}{2023}.
\newblock \bibinfo{title}{National {{Scale Land Cover Classification Using}} the {{Semiautomatic High-Quality Reference Sample Generation}} ({{HRSG}}) {{Method}} and an {{Adaptive Supervised Classification Scheme}}}.
\newblock \bibinfo{journal}{IEEE J. Sel. Top. Appl. Earth Observations Remote Sensing} \bibinfo{volume}{16}, \bibinfo{pages}{1858--1870}.
\newblock \bibinfo{note}{\url{https://ieeexplore.ieee.org/document/10035401/}}.
\bibitem[{Nanni et~al.(2017)Nanni, Ghidoni and Brahnam}]{nanniHandcraftedVsNonhandcrafted2017a}
\bibinfo{author}{Nanni, L.}, \bibinfo{author}{Ghidoni, S.}, \bibinfo{author}{Brahnam, S.}, \bibinfo{year}{2017}.
\newblock \bibinfo{title}{Handcrafted vs. non-handcrafted features for computer vision classification}.
\newblock \bibinfo{journal}{Pattern Recognition} \bibinfo{volume}{71}, \bibinfo{pages}{158--172}.
\newblock \bibinfo{note}{\url{https://linkinghub.elsevier.com/retrieve/pii/S0031320317302224}}.
\bibitem[{North et~al.(2019)North, Pairman and Belliss}]{northBoundaryDelineationAgricultural2019b}
\bibinfo{author}{North, H.C.}, \bibinfo{author}{Pairman, D.}, \bibinfo{author}{Belliss, S.E.}, \bibinfo{year}{2019}.
\newblock \bibinfo{title}{Boundary {{Delineation}} of {{Agricultural Fields}} in {{Multitemporal Satellite Imagery}}}.
\newblock \bibinfo{journal}{IEEE J. Sel. Top. Appl. Earth Observations Remote Sensing} \bibinfo{volume}{12}, \bibinfo{pages}{237--251}.
\newblock \bibinfo{note}{\url{https://ieeexplore.ieee.org/document/8584043/}}.
\bibitem[{Oliphant et~al.(2019)Oliphant, Thenkabail, Teluguntla, Xiong, Gumma, Congalton and Yadav}]{oliphantMappingCroplandExtent2019}
\bibinfo{author}{Oliphant, A.J.}, \bibinfo{author}{Thenkabail, P.S.}, \bibinfo{author}{Teluguntla, P.}, \bibinfo{author}{Xiong, J.}, \bibinfo{author}{Gumma, M.K.}, \bibinfo{author}{Congalton, R.G.}, \bibinfo{author}{Yadav, K.}, \bibinfo{year}{2019}.
\newblock \bibinfo{title}{Mapping cropland extent of {{Southeast}} and {{Northeast Asia}} using multi-year time-series {{Landsat}} 30-m data using a random forest classifier on the {{Google Earth Engine Cloud}}}.
\newblock \bibinfo{journal}{International Journal of Applied Earth Observation and Geoinformation} \bibinfo{volume}{81}, \bibinfo{pages}{110--124}.
\newblock \bibinfo{note}{\url{https://linkinghub.elsevier.com/retrieve/pii/S0303243418307414}}.
\bibitem[{Olofsson et~al.(2014)Olofsson, Foody, Herold, Stehman, Woodcock and Wulder}]{olofsson2014good}
\bibinfo{author}{Olofsson, P.}, \bibinfo{author}{Foody, G.M.}, \bibinfo{author}{Herold, M.}, \bibinfo{author}{Stehman, S.V.}, \bibinfo{author}{Woodcock, C.E.}, \bibinfo{author}{Wulder, M.A.}, \bibinfo{year}{2014}.
\newblock \bibinfo{title}{Good practices for estimating area and assessing accuracy of land change}.
\newblock \bibinfo{journal}{Remote sensing of Environment} \bibinfo{volume}{148}, \bibinfo{pages}{42--57}.
\bibitem[{Pelletier et~al.(2016)Pelletier, Valero, Inglada, Champion and Dedieu}]{pelletierAssessingRobustnessRandom2016a}
\bibinfo{author}{Pelletier, C.}, \bibinfo{author}{Valero, S.}, \bibinfo{author}{Inglada, J.}, \bibinfo{author}{Champion, N.}, \bibinfo{author}{Dedieu, G.}, \bibinfo{year}{2016}.
\newblock \bibinfo{title}{Assessing the robustness of {{Random Forests}} to map land cover with high resolution satellite image time series over large areas}.
\newblock \bibinfo{journal}{Remote Sensing of Environment} \bibinfo{volume}{187}, \bibinfo{pages}{156--168}.
\newblock \bibinfo{note}{\url{https://linkinghub.elsevier.com/retrieve/pii/S0034425716303820}}.
\bibitem[{Persello et~al.(2019)Persello, Tolpekin, Bergado and {de By}}]{perselloDelineationAgriculturalFields2019}
\bibinfo{author}{Persello, C.}, \bibinfo{author}{Tolpekin, V.}, \bibinfo{author}{Bergado, J.}, \bibinfo{author}{{de By}, R.}, \bibinfo{year}{2019}.
\newblock \bibinfo{title}{Delineation of agricultural fields in smallholder farms from satellite images using fully convolutional networks and combinatorial grouping}.
\newblock \bibinfo{journal}{Remote Sensing of Environment} \bibinfo{volume}{231}, \bibinfo{pages}{111253}.
\newblock \bibinfo{note}{\url{https://linkinghub.elsevier.com/retrieve/pii/S003442571930272X}}.
\bibitem[{Prince(2019)}]{princeChallengesRemoteSensing2019a}
\bibinfo{author}{Prince, S.D.}, \bibinfo{year}{2019}.
\newblock \bibinfo{title}{Challenges for remote sensing of the {{Sustainable Development Goal SDG}} 15.3.1 productivity indicator}.
\newblock \bibinfo{journal}{Remote Sensing of Environment} \bibinfo{volume}{234}, \bibinfo{pages}{111428}.
\newblock \bibinfo{note}{\url{https://linkinghub.elsevier.com/retrieve/pii/S003442571930447X}}.
\bibitem[{Rustowicz et~al.(2019)Rustowicz, Cheong, Wang, Ermon, Burke and Lobell}]{Rustowicz2019SemanticSO}
\bibinfo{author}{Rustowicz, R.}, \bibinfo{author}{Cheong, R.}, \bibinfo{author}{Wang, L.}, \bibinfo{author}{Ermon, S.}, \bibinfo{author}{Burke, M.}, \bibinfo{author}{Lobell, D.}, \bibinfo{year}{2019}.
\newblock \bibinfo{title}{Semantic segmentation of crop type in africa: A novel dataset and analysis of deep learning methods}, in: \bibinfo{booktitle}{CVPR Workshops}, pp. \bibinfo{pages}{75--82}.
\newblock \bibinfo{note}{\url{https://api.semanticscholar.org/CorpusID:198180478}}.
\bibitem[{Sabokrou et~al.(2019)Sabokrou, Khalooei and Adeli}]{sabokrouSelfSupervisedRepresentationLearning2019a}
\bibinfo{author}{Sabokrou, M.}, \bibinfo{author}{Khalooei, M.}, \bibinfo{author}{Adeli, E.}, \bibinfo{year}{2019}.
\newblock \bibinfo{title}{Self-{{Supervised Representation Learning}} via {{Neighborhood-Relational Encoding}}}, in: \bibinfo{booktitle}{2019 {{IEEECVF Int}}. {{Conf}}. {{Comput}}. {{Vis}}. {{ICCV}}}, \bibinfo{publisher}{{IEEE}}, \bibinfo{address}{{Seoul, Korea (South)}}. pp. \bibinfo{pages}{8009--8018}.
\newblock \bibinfo{note}{\url{https://ieeexplore.ieee.org/document/9010354/}}.
\bibitem[{Shi et~al.(2015)Shi, Chen, Wang, Yeung, Wong and Woo}]{shi2015convolutional}
\bibinfo{author}{Shi, X.}, \bibinfo{author}{Chen, Z.}, \bibinfo{author}{Wang, H.}, \bibinfo{author}{Yeung, D.Y.}, \bibinfo{author}{Wong, W.K.}, \bibinfo{author}{Woo, W.c.}, \bibinfo{year}{2015}.
\newblock \bibinfo{title}{Convolutional lstm network: A machine learning approach for precipitation nowcasting}.
\newblock \bibinfo{journal}{Advances in neural information processing systems} \bibinfo{volume}{28}.
\newblock \bibinfo{note}{\url{https://proceedings.neurips.cc/paper_files/paper/2015/file/07563a3fe3bbe7e3ba84431ad9d055af-Paper.pdf}}.
\bibitem[{Singh et~al.(2022)Singh, Singh, Sethi and Sood}]{singhDeepLearningMapping2022a}
\bibinfo{author}{Singh, G.}, \bibinfo{author}{Singh, S.}, \bibinfo{author}{Sethi, G.}, \bibinfo{author}{Sood, V.}, \bibinfo{year}{2022}.
\newblock \bibinfo{title}{Deep {{Learning}} in the {{Mapping}} of {{Agricultural Land Use Using Sentinel-2 Satellite Data}}}.
\newblock \bibinfo{journal}{Geographies} \bibinfo{volume}{2}, \bibinfo{pages}{691--700}.
\newblock \bibinfo{note}{\url{https://www.mdpi.com/2673-7086/2/4/42}}.
\bibitem[{Stinson et~al.(2016)Stinson, Magnussen, Boudewyn, Eichel, Russo, Cranny and Song}]{stinsonCanada2016}
\bibinfo{author}{Stinson, G.}, \bibinfo{author}{Magnussen, S.}, \bibinfo{author}{Boudewyn, P.}, \bibinfo{author}{Eichel, F.}, \bibinfo{author}{Russo, G.}, \bibinfo{author}{Cranny, M.}, \bibinfo{author}{Song, A.}, \bibinfo{year}{2016}.
\newblock \bibinfo{title}{Canada}, in: \bibinfo{editor}{Vidal, C.}, \bibinfo{editor}{Alberdi, I.A.}, \bibinfo{editor}{Hern{\'a}ndez~Mateo, L.}, \bibinfo{editor}{Redmond, J.J.} (Eds.), \bibinfo{booktitle}{National {{Forest Inventories}}}. \bibinfo{publisher}{{Springer International Publishing}}, \bibinfo{address}{{Cham}}, pp. \bibinfo{pages}{233--247}.
\newblock \bibinfo{note}{\url{http://link.springer.com/10.1007/978-3-319-44015-6_12}}.
\bibitem[{Sun et~al.(2019)Sun, Di and Fang}]{sunUsingLongShortterm2019a}
\bibinfo{author}{Sun, Z.}, \bibinfo{author}{Di, L.}, \bibinfo{author}{Fang, H.}, \bibinfo{year}{2019}.
\newblock \bibinfo{title}{Using long short-term memory recurrent neural network in land cover classification on {{Landsat}} and {{Cropland}} data layer time series}.
\newblock \bibinfo{journal}{International Journal of Remote Sensing} \bibinfo{volume}{40}, \bibinfo{pages}{593--614}.
\newblock \bibinfo{note}{\url{https://www.tandfonline.com/doi/full/10.1080/01431161.2018.1516313}}.
\bibitem[{Sykas et~al.(2022)Sykas, Sdraka, Zografakis and Papoutsis}]{sykasSentinel2MultiyearMulticountry2022b}
\bibinfo{author}{Sykas, D.}, \bibinfo{author}{Sdraka, M.}, \bibinfo{author}{Zografakis, D.}, \bibinfo{author}{Papoutsis, I.}, \bibinfo{year}{2022}.
\newblock \bibinfo{title}{A {{Sentinel-2 Multiyear}}, {{Multicountry Benchmark Dataset}} for {{Crop Classification}} and {{Segmentation With Deep Learning}}}.
\newblock \bibinfo{journal}{IEEE J. Sel. Top. Appl. Earth Observations Remote Sensing} \bibinfo{volume}{15}, \bibinfo{pages}{3323--3339}.
\newblock \bibinfo{note}{\url{https://ieeexplore.ieee.org/document/9749916/}}.
\bibitem[{Wagner and Oppelt(2020)}]{wagnerExtractingAgriculturalFields2020a}
\bibinfo{author}{Wagner, M.P.}, \bibinfo{author}{Oppelt, N.}, \bibinfo{year}{2020}.
\newblock \bibinfo{title}{Extracting {{Agricultural Fields}} from {{Remote Sensing Imagery Using Graph-Based Growing Contours}}}.
\newblock \bibinfo{journal}{Remote Sensing} \bibinfo{volume}{12}, \bibinfo{pages}{1205}.
\newblock \bibinfo{note}{\url{https://www.mdpi.com/2072-4292/12/7/1205}}.
\bibitem[{Wang et~al.(2022)Wang, Wang, Shen, Fei, Li, Jin, Wu, Zhao and Le}]{wangSemiSupervisedSemanticSegmentation2022b}
\bibinfo{author}{Wang, Y.}, \bibinfo{author}{Wang, H.}, \bibinfo{author}{Shen, Y.}, \bibinfo{author}{Fei, J.}, \bibinfo{author}{Li, W.}, \bibinfo{author}{Jin, G.}, \bibinfo{author}{Wu, L.}, \bibinfo{author}{Zhao, R.}, \bibinfo{author}{Le, X.}, \bibinfo{year}{2022}.
\newblock \bibinfo{title}{Semi-{{Supervised Semantic Segmentation Using Unreliable Pseudo-Labels}}}, in: \bibinfo{booktitle}{2022 {{IEEECVF Conf}}. {{Comput}}. {{Vis}}. {{Pattern Recognit}}. {{CVPR}}}, \bibinfo{publisher}{{IEEE}}, \bibinfo{address}{{New Orleans, LA, USA}}. pp. \bibinfo{pages}{4238--4247}.
\newblock \bibinfo{note}{\url{https://ieeexplore.ieee.org/document/9879387/}}.
\bibitem[{Weiss et~al.(2020)Weiss, Jacob and Duveiller}]{weissRemoteSensingAgricultural2020a}
\bibinfo{author}{Weiss, M.}, \bibinfo{author}{Jacob, F.}, \bibinfo{author}{Duveiller, G.}, \bibinfo{year}{2020}.
\newblock \bibinfo{title}{Remote sensing for agricultural applications: {{A}} meta-review}.
\newblock \bibinfo{journal}{Remote Sensing of Environment} \bibinfo{volume}{236}, \bibinfo{pages}{111402}.
\newblock \bibinfo{note}{\url{https://linkinghub.elsevier.com/retrieve/pii/S0034425719304213}}.
\bibitem[{Wulder et~al.(2018)Wulder, Li, Campbell, White, Hobart, Hermosilla and Coops}]{wulderNationalAssessmentWetland2018}
\bibinfo{author}{Wulder, M.}, \bibinfo{author}{Li, Z.}, \bibinfo{author}{Campbell, E.}, \bibinfo{author}{White, J.}, \bibinfo{author}{Hobart, G.}, \bibinfo{author}{Hermosilla, T.}, \bibinfo{author}{Coops, N.}, \bibinfo{year}{2018}.
\newblock \bibinfo{title}{A {{National Assessment}} of {{Wetland Status}} and {{Trends}} for {{Canada}}'s {{Forested Ecosystems Using}} 33 {{Years}} of {{Earth Observation Satellite Data}}}.
\newblock \bibinfo{journal}{Remote Sensing} \bibinfo{volume}{10}, \bibinfo{pages}{1623}.
\newblock \bibinfo{note}{\url{http://www.mdpi.com/2072-4292/10/10/1623}}.
\bibitem[{Wulder et~al.(2003)Wulder, Dechka, Gillis, Luther, Hall, Beaudoin and Franklin}]{wulderOperationalMappingLand2003}
\bibinfo{author}{Wulder, M.A.}, \bibinfo{author}{Dechka, J.A.}, \bibinfo{author}{Gillis, M.A.}, \bibinfo{author}{Luther, J.E.}, \bibinfo{author}{Hall, R.J.}, \bibinfo{author}{Beaudoin, A.}, \bibinfo{author}{Franklin, S.E.}, \bibinfo{year}{2003}.
\newblock \bibinfo{title}{Operational mapping of the land cover of the forested area of {{Canada}} with {{Landsat}} data: {{EOSD}} land cover program}.
\newblock \bibinfo{journal}{The Forestry Chronicle} \bibinfo{volume}{79}, \bibinfo{pages}{1075--1083}.
\newblock \bibinfo{note}{\url{http://pubs.cif-ifc.org/doi/10.5558/tfc791075-6}}.
\bibitem[{Xu et~al.(2024)Xu, Yao, Zhang, Yang, Feng, Li, Yan, Gao, Li, Yang et~al.}]{xu2024deep}
\bibinfo{author}{Xu, F.}, \bibinfo{author}{Yao, X.}, \bibinfo{author}{Zhang, K.}, \bibinfo{author}{Yang, H.}, \bibinfo{author}{Feng, Q.}, \bibinfo{author}{Li, Y.}, \bibinfo{author}{Yan, S.}, \bibinfo{author}{Gao, B.}, \bibinfo{author}{Li, S.}, \bibinfo{author}{Yang, J.}, et~al., \bibinfo{year}{2024}.
\newblock \bibinfo{title}{Deep learning in cropland field identification: A review}.
\newblock \bibinfo{journal}{Computers and Electronics in Agriculture} \bibinfo{volume}{222}, \bibinfo{pages}{109042}.
\bibitem[{Xu et~al.(2017)Xu, Yu, Zhao, Feng, Cheng, Cai and Gong}]{xuMonitoringCroplandChanges2017a}
\bibinfo{author}{Xu, Y.}, \bibinfo{author}{Yu, L.}, \bibinfo{author}{Zhao, Y.}, \bibinfo{author}{Feng, D.}, \bibinfo{author}{Cheng, Y.}, \bibinfo{author}{Cai, X.}, \bibinfo{author}{Gong, P.}, \bibinfo{year}{2017}.
\newblock \bibinfo{title}{Monitoring cropland changes along the {{Nile River}} in {{Egypt}} over past three decades (1984{\textendash}2015) using remote sensing}.
\newblock \bibinfo{journal}{International Journal of Remote Sensing} \bibinfo{volume}{38}, \bibinfo{pages}{4459--4480}.
\newblock \bibinfo{note}{\url{https://www.tandfonline.com/doi/full/10.1080/01431161.2017.1323285}}.
\bibitem[{Yin et~al.(2021)Yin, Dong, Hamm, Li, Wang, Xing and Fu}]{yinIntegratingRemoteSensing2021b}
\bibinfo{author}{Yin, J.}, \bibinfo{author}{Dong, J.}, \bibinfo{author}{Hamm, N.A.}, \bibinfo{author}{Li, Z.}, \bibinfo{author}{Wang, J.}, \bibinfo{author}{Xing, H.}, \bibinfo{author}{Fu, P.}, \bibinfo{year}{2021}.
\newblock \bibinfo{title}{Integrating remote sensing and geospatial big data for urban land use mapping: {{A}} review}.
\newblock \bibinfo{journal}{International Journal of Applied Earth Observation and Geoinformation} \bibinfo{volume}{103}, \bibinfo{pages}{102514}.
\newblock \bibinfo{note}{\url{https://linkinghub.elsevier.com/retrieve/pii/S030324342100221X}}.
\bibitem[{Yu et~al.(2013)Yu, Wang and Gong}]{yuImproving30Global2013a}
\bibinfo{author}{Yu, L.}, \bibinfo{author}{Wang, J.}, \bibinfo{author}{Gong, P.}, \bibinfo{year}{2013}.
\newblock \bibinfo{title}{Improving 30 m global land-cover map {{FROM-GLC}} with time series {{MODIS}} and auxiliary data sets: A segmentation-based approach}.
\newblock \bibinfo{journal}{International Journal of Remote Sensing} \bibinfo{volume}{34}, \bibinfo{pages}{5851--5867}.
\newblock \bibinfo{note}{\url{https://www.tandfonline.com/doi/full/10.1080/01431161.2013.798055}}.
\bibitem[{Yue et~al.(2013)Yue, Zhang, Yang, Su, Yun and Zhu}]{yueTextureExtractionObjectoriented2013}
\bibinfo{author}{Yue, A.}, \bibinfo{author}{Zhang, C.}, \bibinfo{author}{Yang, J.}, \bibinfo{author}{Su, W.}, \bibinfo{author}{Yun, W.}, \bibinfo{author}{Zhu, D.}, \bibinfo{year}{2013}.
\newblock \bibinfo{title}{Texture extraction for object-oriented classification of high spatial resolution remotely sensed images using a semivariogram}.
\newblock \bibinfo{journal}{International Journal of Remote Sensing} \bibinfo{volume}{34}, \bibinfo{pages}{3736--3759}.
\newblock \bibinfo{note}{\url{https://www.tandfonline.com/doi/full/10.1080/01431161.2012.759298}}.
\bibitem[{Zanaga et~al.(2022a)Zanaga, Van De~Kerchove, Daems, De~Keersmaecker, Brockmann, Kirches, Wevers, Cartus, Santoro, Fritz, Lesiv, Herold, Tsendbazar, Xu, Ramoino and Arino}]{zanagaESAWorldCover102022}
\bibinfo{author}{Zanaga, D.}, \bibinfo{author}{Van De~Kerchove, R.}, \bibinfo{author}{Daems, D.}, \bibinfo{author}{De~Keersmaecker, W.}, \bibinfo{author}{Brockmann, C.}, \bibinfo{author}{Kirches, G.}, \bibinfo{author}{Wevers, J.}, \bibinfo{author}{Cartus, O.}, \bibinfo{author}{Santoro, M.}, \bibinfo{author}{Fritz, S.}, \bibinfo{author}{Lesiv, M.}, \bibinfo{author}{Herold, M.}, \bibinfo{author}{Tsendbazar, N.E.}, \bibinfo{author}{Xu, P.}, \bibinfo{author}{Ramoino, F.}, \bibinfo{author}{Arino, O.}, \bibinfo{year}{2022}a.
\newblock \bibinfo{title}{{{ESA WorldCover}} 10 m 2021 v200}.
\newblock \bibinfo{howpublished}{\url{https://zenodo.org/record/7254220}}.
\bibitem[{Zanaga et~al.(2022b)Zanaga, Van De~Kerchove, Daems, De~Keersmaecker, Brockmann, Kirches, Wevers, Cartus, Santoro, Fritz et~al.}]{zanaga2022esa}
\bibinfo{author}{Zanaga, D.}, \bibinfo{author}{Van De~Kerchove, R.}, \bibinfo{author}{Daems, D.}, \bibinfo{author}{De~Keersmaecker, W.}, \bibinfo{author}{Brockmann, C.}, \bibinfo{author}{Kirches, G.}, \bibinfo{author}{Wevers, J.}, \bibinfo{author}{Cartus, O.}, \bibinfo{author}{Santoro, M.}, \bibinfo{author}{Fritz, S.}, et~al., \bibinfo{year}{2022}b.
\newblock \bibinfo{title}{Esa worldcover 10 m 2021 v200} .
\bibitem[{Zhang et~al.(2020)Zhang, Pan, Zhang, Hu, Zhao, Li and Chen}]{zhangGeneralizedApproachBased2020e}
\bibinfo{author}{Zhang, D.}, \bibinfo{author}{Pan, Y.}, \bibinfo{author}{Zhang, J.}, \bibinfo{author}{Hu, T.}, \bibinfo{author}{Zhao, J.}, \bibinfo{author}{Li, N.}, \bibinfo{author}{Chen, Q.}, \bibinfo{year}{2020}.
\newblock \bibinfo{title}{A generalized approach based on convolutional neural networks for large area cropland mapping at very high resolution}.
\newblock \bibinfo{journal}{Remote Sensing of Environment} \bibinfo{volume}{247}, \bibinfo{pages}{111912}.
\newblock \bibinfo{note}{\url{https://linkinghub.elsevier.com/retrieve/pii/S0034425720302820}}.
\bibitem[{Zhang and Roy(2017)}]{zhangUsing500MODIS2017b}
\bibinfo{author}{Zhang, H.K.}, \bibinfo{author}{Roy, D.P.}, \bibinfo{year}{2017}.
\newblock \bibinfo{title}{Using the 500 m {{MODIS}} land cover product to derive a consistent continental scale 30 m {{Landsat}} land cover classification}.
\newblock \bibinfo{journal}{Remote Sensing of Environment} \bibinfo{volume}{197}, \bibinfo{pages}{15--34}.
\newblock \bibinfo{note}{\url{https://linkinghub.elsevier.com/retrieve/pii/S0034425717302249}}.
\bibitem[{Zhang et~al.(2023)Zhang, Guo, Zhang, Xia, Zhang, Lin, Tang, Fang and Du}]{zhangNovelKnowledgeDrivenAutomated2023a}
\bibinfo{author}{Zhang, W.}, \bibinfo{author}{Guo, S.}, \bibinfo{author}{Zhang, P.}, \bibinfo{author}{Xia, Z.}, \bibinfo{author}{Zhang, X.}, \bibinfo{author}{Lin, C.}, \bibinfo{author}{Tang, P.}, \bibinfo{author}{Fang, H.}, \bibinfo{author}{Du, P.}, \bibinfo{year}{2023}.
\newblock \bibinfo{title}{A {{Novel Knowledge-Driven Automated Solution}} for {{High-Resolution Cropland Extraction}} by {{Cross-Scale Sample Transfer}}}.
\newblock \bibinfo{journal}{IEEE Trans. Geosci. Remote Sensing} \bibinfo{volume}{61}, \bibinfo{pages}{1--16}.
\newblock \bibinfo{note}{\url{https://ieeexplore.ieee.org/document/10197441/}}.
\bibitem[{Zhong et~al.(2019)Zhong, Hu and Zhou}]{zhong2019deep}
\bibinfo{author}{Zhong, L.}, \bibinfo{author}{Hu, L.}, \bibinfo{author}{Zhou, H.}, \bibinfo{year}{2019}.
\newblock \bibinfo{title}{Deep learning based multi-temporal crop classification}.
\newblock \bibinfo{journal}{Remote sensing of environment} \bibinfo{volume}{221}, \bibinfo{pages}{430--443}.
\bibitem[{Zhu et~al.(2017)Zhu, Tuia, Mou, Xia, Zhang, Xu and Fraundorfer}]{zhuDeepLearningRemote2017}
\bibinfo{author}{Zhu, X.X.}, \bibinfo{author}{Tuia, D.}, \bibinfo{author}{Mou, L.}, \bibinfo{author}{Xia, G.S.}, \bibinfo{author}{Zhang, L.}, \bibinfo{author}{Xu, F.}, \bibinfo{author}{Fraundorfer, F.}, \bibinfo{year}{2017}.
\newblock \bibinfo{title}{Deep {{Learning}} in {{Remote Sensing}}: {{A Comprehensive Review}} and {{List}} of {{Resources}}}.
\newblock \bibinfo{journal}{IEEE Geosci. Remote Sens. Mag.} \bibinfo{volume}{5}, \bibinfo{pages}{8--36}.
\newblock \DOIprefix\doi{10.1109/MGRS.2017.2762307}.
\bibitem[{Zhu et~al.(2016)Zhu, Gallant, Woodcock, Pengra, Olofsson, Loveland, Jin, Dahal, Yang and Auch}]{zhuOptimizingSelectionTraining2016a}
\bibinfo{author}{Zhu, Z.}, \bibinfo{author}{Gallant, A.L.}, \bibinfo{author}{Woodcock, C.E.}, \bibinfo{author}{Pengra, B.}, \bibinfo{author}{Olofsson, P.}, \bibinfo{author}{Loveland, T.R.}, \bibinfo{author}{Jin, S.}, \bibinfo{author}{Dahal, D.}, \bibinfo{author}{Yang, L.}, \bibinfo{author}{Auch, R.F.}, \bibinfo{year}{2016}.
\newblock \bibinfo{title}{Optimizing selection of training and auxiliary data for operational land cover classification for the {{LCMAP}} initiative}.
\newblock \bibinfo{journal}{ISPRS Journal of Photogrammetry and Remote Sensing} \bibinfo{volume}{122}, \bibinfo{pages}{206--221}.
\newblock \bibinfo{note}{\url{https://linkinghub.elsevier.com/retrieve/pii/S0924271616302829}}.

\end{thebibliography}




\end{document}